%% file: main.tex
\documentclass[lettersize,journal]{IEEEtran}
\usepackage{amsmath,amsfonts}
\usepackage{algorithmic}
\usepackage{algorithm}
\usepackage{array}
\usepackage[caption=false,font=normalsize,labelfont=sf,textfont=sf]{subfig}
\usepackage{textcomp}
\usepackage{stfloats}
\usepackage{url}
\usepackage{verbatim}
\usepackage{graphicx}
\usepackage{cite}

\usepackage{amsthm,amssymb,amsfonts}
\usepackage{booktabs}
\usepackage{multirow}
\usepackage{pifont}
\usepackage[table]{xcolor}
\definecolor{darkgreen}{rgb}{0, 0.5, 0}
\usepackage{svg}
\usepackage{ulem}

\begin{document}

\title{DualDiff+: Dual-Branch Diffusion for High-Fidelity Video Generation with Reward Guidance}

\author{Zhao Yang, Zezhong Qian, Xiaofan Li, Weixiang Xu~\IEEEmembership{Member,~IEEE},\\ Gongpeng Zhao, Ruohong Yu, Lingsi Zhu and Longjun Liu,~\IEEEmembership{Member,~IEEE}
\thanks{The first two authors contributed equally.}
\thanks{Zhao Yang, Zezhong Qian, Ruohong Yu and Longjun Liu are with the
National Key Laboratory of Human-Machine Hybrid Augmented Intelligence,National Engineering Research Center for Visual Information and Applications, and Institute of Artificial Intelligence and Robotics, Xi’an Jiaotong University. (e-mail: {yangzhao17; zezhongqian; 2233113339}@stu.xjtu.edu.cn; liulongjun@xjtu.edu.cn)}
\thanks{Xiaofan Li is with the College of Optical Science and Engineering, Zhejiang University, Hangzhou 310027, China. (e-mail: shalfunnn@gmail.com)}
\thanks{Weixiang Xu is with Institute of Automation, Chinese Academy of Sciences, Beijing 100190, China. (e-mail: wxxu218@gmail.com)}
\thanks{Lingsi Zhu and Gongpeng Zhao are with the University of Science and Technology of China, Anhui 230052, China. (e-mail: {ls-zhu24, zgp0531}@mail.ustc.edu.cn)}
}

\markboth{Journal of \LaTeX\ Class Files,~Vol.~14, No.~8, August~2021}%
{Shell \MakeLowercase{\textit{et al.}}: A Sample Article Using IEEEtran.cls for IEEE Journals}



\maketitle

\begin{abstract}
Accurate and high-fidelity driving scene reconstruction demands the effective utilization of comprehensive scene information as conditional inputs. Existing methods predominantly rely on 3D bounding boxes and BEV road maps for foreground and background control, which fail to capture the full complexity of driving scenes and adequately integrate multimodal information. In this work, we present DualDiff, a dual-branch conditional diffusion model designed to enhance driving scene generation across multiple views and video sequences. Specifically, we introduce Occupancy Ray-shape Sampling (ORS) as a conditional input, offering rich foreground and background semantics alongside 3D spatial geometry to precisely control the generation of both elements. To improve the synthesis of fine-grained foreground objects, particularly complex and distant ones, we propose a Foreground-Aware Mask (FGM) denoising loss function. Additionally, we develop the Semantic Fusion Attention (SFA) mechanism to dynamically prioritize relevant information and suppress noise, enabling more effective multimodal fusion. Finally, to ensure high-quality image-to-video generation, we introduce the Reward-Guided Diffusion (RGD) framework, which maintains global consistency and semantic coherence in generated videos. Extensive experiments demonstrate that DualDiff achieves state-of-the-art (SOTA) performance across multiple datasets. On the NuScenes dataset, DualDiff reduces the FID score by 4.09\% compared to the best baseline. In downstream tasks, such as BEV segmentation, our method improves vehicle mIoU by 4.50\% and road mIoU by 1.70\%, while in BEV 3D object detection, the foreground mAP increases by 1.46\%. Code will be made available at \url{https://github.com/yangzhaojason/DualDiff}.
\end{abstract}



\input{sec/1_introduction}
\input{sec/2_related}
\input{sec/3_method}
\input{sec/4_exp}

\section{Conclusion}
Accurate and high-fidelity driving scene generation is crucial for autonomous perception, simulation, and planning. In this work, we introduced DualDiff, a dual-branch conditional diffusion model for multi-view and video-based scene synthesis. By leveraging Occupancy Ray-shape Sampling (ORS) for enriched 3D spatial semantics, Foreground-Aware Mask (FGM) loss for fine-grained object synthesis, and Semantic Fusion Attention (SFA) for multimodal fusion, our method enhances fidelity and controllability. Additionally, the Reward-Guided Diffusion (RGD) framework ensures coherent and semantically accurate image-to-video generation. Extensive experiments demonstrate that DualDiff achieves state-of-the-art performance across multiple datasets, effectively generating realistic, geometry-aware scenes with precise foreground and background control. Future work will focus on improving computational efficiency, expanding generalization to diverse environments, and integrating multi-sensor fusion for enhanced robustness in autonomous driving applications.





%

\bibliographystyle{IEEEtran}
\normalem
\bibliography{main}












\newpage

 





\end{document}

%% file: sec/1_introduction.tex
\section{Introduction}
\IEEEPARstart{M}ost autonomous driving research relies on large-scale camera datasets with detailed annotations~\cite{cui2024survey}, \cite{MARTINEZDIAZ2018275}, \cite{chen2024end}. However, the high cost of data collection and annotation limits the availability of open-source vision datasets~\cite{BOSCH201876}, \cite{10432820}. Advanced generative models, such as Stable Diffusion~\cite{rombach2022high}, \cite{li2023drivingdiffusion}, \cite{song2020denoising}, offer a promising alternative by generating realistic images for synthesizing street-view data. These models reduce the reliance on expensive real-world data while enabling the creation of diverse and high-quality synthetic datasets.

Conditional generative models for autonomous driving have made significant progress in generating high-fidelity scenes, which are valuable for downstream vision tasks. However, several limitations persist:
1) \textbf{Suboptimal condition encoding method}: Existing methods~\cite{wen2024panacea, gao2023magicdrive, zhang2024perldiff} often rely on 3D vectorized inputs or BEV layouts to represent obstacles and lane markings. However, these approaches face key limitations. They misalign with Perspective View (PV) encoding, causing mismatches with perspective-based image generation models. Additionally, 3D bounding boxes lack fine details and do not align with the conditioning format of pretrained generative models like \textit{Stable Diffusion}. These issues highlight the need for a unified representation that better integrates 3D scene understanding with image generation. Furthermore, treating foreground and background as the same input may limit the model's learning capacity.
2) \textbf{Insufficient cross-modal interaction}: Current methods~\cite{gao2023magicdrive, wang2024drivewm} fuse multimodal inputs (e.g., maps, bounding boxes, prompts) using direct concatenation or static attention, lacking dynamic adaptation to prioritize useful information or suppress noise. This results in suboptimal feature integration, compromising scene generation quality. A more adaptive fusion mechanism is needed to enhance cross-modal interaction and improve output coherence.
3) \textbf{Lack of overall consistency and coherence in image-to-video generation}: While existing methods focus on pixel-level details, they often neglect the global consistency and instance coherence of the generated videos. This limitation makes it challenging to ensure that the generated videos meet high-level perceptual quality or task-specific requirements.

In this paper, we introduce DualDiff, a dual-branch architecture that offers comprehensive scene control for high-quality image and video generation. We address key challenges with the following innovations: 
1) \textbf{Comprehensive scene control with a dual-branch architecture}: We propose an Occupancy Ray-shape Sampling (ORS) method, which offers rich detail and more precise control as multi-branch conditional inputs. In addition, we introduce a Foreground-Aware Mask (FGM) loss, which applies a weighted mask to the original diffusion model denoising loss, effectively improving the generation of distant fine-grained objects. 2) \textbf{Cross-modal semantic interaction}: To address the alignment of information across diverse modalities, we introduce a Semantic Fusion Attention (SFA) mechanism. This algorithm dynamically integrates multimodal inputs by adaptively focusing on salient features while filtering out noise, thereby significantly enhancing the robustness and precision of cross-modal information fusion. 3) \textbf{Enhanced video generation with advanced semantic understanding}: We introduce a novel Reward-Guided Diffusion (RGD) framework, which incorporates a reward-driven alignment mechanism to enhance the overall quality and applicability of video-generated driving scenarios. These contributions enable DualDiff to generate high-quality, semantically coherent, and contextually accurate driving scenarios for both image and video generation tasks.

Our proposed model effectively integrates cross-modal scene features, enabling accurate reconstruction of scene content, and outperforms existing methods in terms of image and video style fidelity, foreground attributes, and background layout accuracy. The main contributions of this work are summarized as follows:
\begin{itemize}
\item
We propose a novel dual-branch design that integrates Occupancy Ray-shape Sampling (ORS) and Foreground-Aware Mask (FGM) Loss. ORS captures detailed semantics and 3D spatial geometry information, while FGM enhances the generation of distant fine-grained objects, achieving state-of-the-art (SOTA) performance.
\item
We introduce a semantic fusion attention (SFA) mechanism that unifies spatial physics, textual semantics, and dense 3D visual features, dynamically selecting important information and filtering out noise, ensuring seamless alignment of multimodal inputs.
\item 
We present the Reward-Guided Diffusion (RGD) framework, which incorporates a reward-driven alignment strategy to refine video generation, improving the overall quality, consistency, and task-specific relevance of generated driving scenarios.
\end{itemize}

This work is the expanded version of our previous research \cite{dualdiff2025}. 1) Theoretically, we introduce the Reward-Guided Diffusion (RGD) framework, which integrates diffusion models with reward-guided mechanisms by computing rewards from end-to-end inference results in Section~\ref{method:reward-guided}. 2) Experimentally, we extend our comparative analysis in Section~\ref{ex:main_pq}, focusing on the effectiveness of the DualDiff structure rather than the improvement brought by increasing model parameters. In Section~\ref{ex:closed_loop}, we evaluate the role of synthetic data in data-centric closed-loop systems, emphasizing its importance in multi-stage training. Finally, we compare the video metric of many video generation methods (Table~\ref{tab:fvd}) and analyze the significant improvement brought by RGD (Section~\ref{ex:main_video_quality}).

%% file: sec/2_related.tex
\section{Related work}
\label{sec:rwork}
\subsection{Diffusion Models for Conditional Generation}
Recent advances in conditional diffusion models have introduced techniques for generating realistic, contextually rich content. Methods like ControlNet~\cite{zhang2023controlnet} and UniPC~\cite{zhao2024uni} integrate external networks to provide fine-grained control over outputs. Other models, such as Drive-WM~\cite{wang2024drivewm} and Disco~\cite{wang2023disco}, use cross-attention mechanisms within the UNet architecture, while others ~\cite{Singer2022makeavideo}, \cite{he2023animate} incorporate conditions via element-wise operations with noise. In autonomous driving, BEVControl~\cite{yang2023bevcontrol} combines bird's-eye view and street view data for geometrically consistent foreground generation, and MagicDrive~\cite{gao2023magicdrive} integrates BEV maps, 3D bounding boxes, and camera poses for detailed 3D spatial information. DrivingDiffusion~\cite{li2023drivingdiffusion} ensures view coherence through consistency loss, while Panacea~\cite{wen2024panacea} emphasizes temporal stability. PerlDiff~\cite{zhang2024perldiff} enhances object control in street view generation using 3D geometric priors. However, these methods often focus on foreground generation, neglecting the importance of background information and fine-grained object generation. Our approach addresses these limitations by integrating both foreground and background information as conditional inputs for more comprehensive content generation.
\subsection{Video Generation for Autonomous Driving}
Recent advances in video generation technology for autonomous driving have significantly leveraged generative models to simulate dynamic driving environments, which are crucial for the testing and development of autonomous systems. Conditional generative models, particularly Generative Adversarial Networks (GANs) and diffusion models, have shown exceptional efficacy in generating realistic driving scenarios. For instance, Panacea~\cite{wen2024panacea} employs a diffusion-based framework to produce high-fidelity driving videos, conditioned on road maps, vehicle states, and sensor data, achieving highly realistic simulations of dynamic driving environments. Similarly, Drive-WM~\cite{wang2024drivewm} focuses on generating realistic vehicle trajectories over time, facilitating future state predictions and supporting decision-making processes in autonomous driving. Additionally, models like DriveDreamer~\cite{wang2023drivedreamer} and DriveDreamer-2~\cite{zhao2024drivedreamer2} advance video generation by incorporating environmental factors to accurately simulate driving behaviors under varying conditions such as different weather and lighting. These advancements have significantly propelled the field of autonomous driving, providing critical tools for testing, validation, and safety assessments in simulated environments. However, these methods primarily focus on pixel-level details, often neglecting the global consistency and semantic coherence of the generated videos, which makes it challenging to ensure that the generated videos meet high-level perceptual quality or task-specific requirements. 

\begin{figure*}
\centering{}
\includegraphics[width=\linewidth]{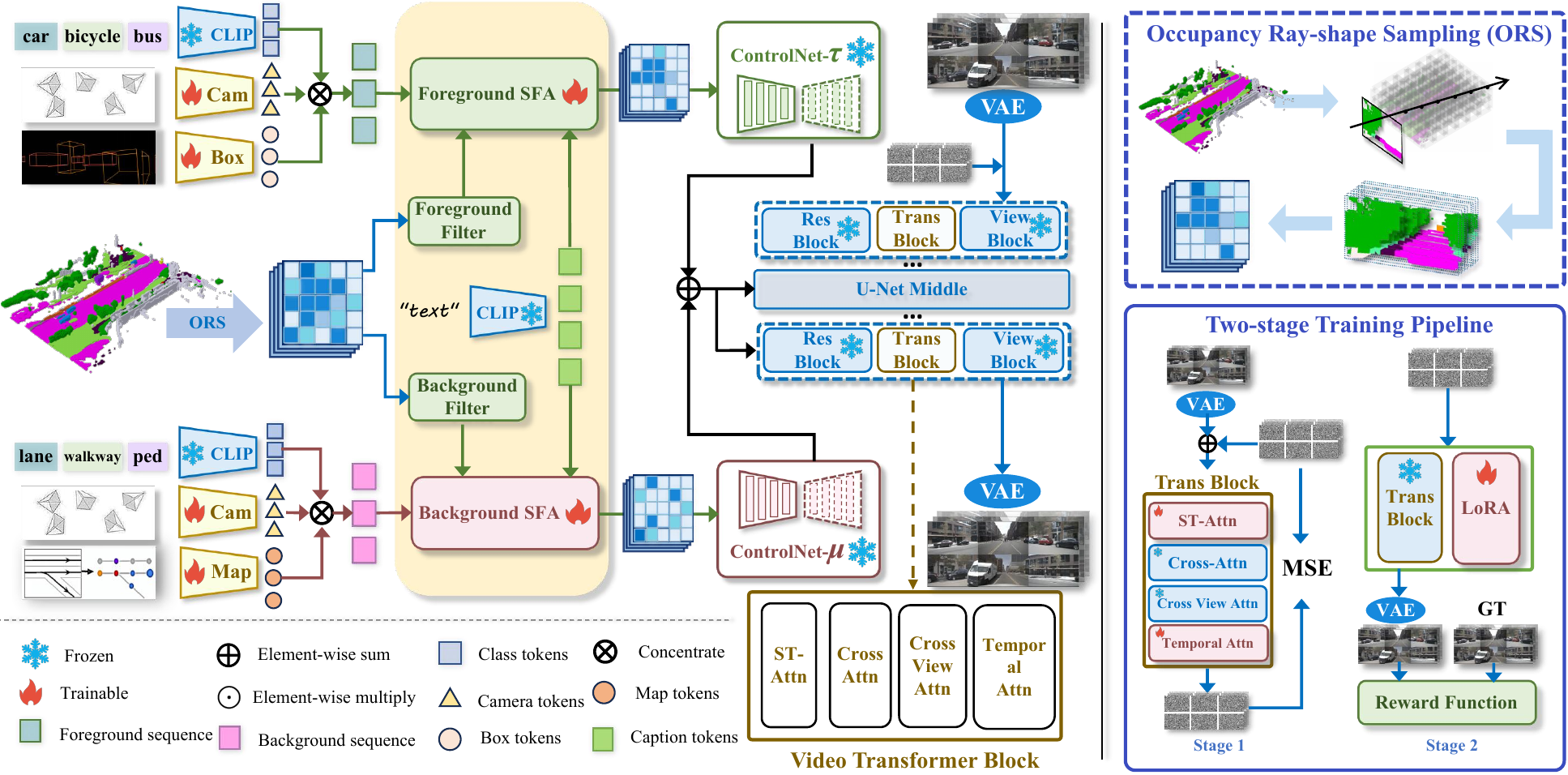}
\caption{\textbf{Architecture Overview of DualDiff for Video Generation}. The model uses Occupancy Ray-shape Sampling (ORS) and Semantic Fusion Attention (SFA) for scene representation, which are fed into a dual-branch foreground-background architecture. The outputs are merged through residual connections in a U-Net. Video generation follows a Two-stage optimization: Spatio-Temporal Attention (ST-Attn) and Temporal Attention (Temporal Attn) are trained in the first stage, while Reward-Guided Diffusion (RGD) and Low-Rank Adaptation (LoRA) fine-tune the attention in the second stage.}
\label{fig:main}
\vspace{-0.2cm}
\end{figure*}

%% file: sec/3_method.tex
\section{Method}
\label{sec:method}

\subsection{Preliminary}
The occupancy grid map is denoted as \( \mathcal{G} \in \mathbb{R}^{H \times W \times D} \), where \( H \), \( W \), and \( D \) represent the height, width, and depth of the grid, respectively. The image dimensions in pixel space are given by height \( U \) and width \( V \). The occupancy grid contains \( N_{\text{sample}} \) ray-sampled points, and \( T \) represents the total number of diffusion timesteps. Our model takes multimodal input conditions, including the occupancy grid map \( \mathcal{G} \), camera pose \( \mathbf{P} \), 3D bounding box \( \mathbf{B} \), vectorized map \( \mathbf{H} \), and textual sequence \( \mathbf{L} \). The output consists of multi-view images and videos generated by the model.

\subsection{Overall Architecture}
We propose a dual-branch conditional control structure, as illustrated in Fig. \ref{fig:main}. The foreground and background branches extract semantic information from the 3D occupancy grid using the Occupancy Ray-shape Sampling (ORS) method. This is followed by multimodal fusion via the Semantic Fusion Attention (SFA) algorithm, combining visual, spatial, and textual data to capture the complex dynamics of autonomous driving scenes. To enhance foreground object generation quality, we introduce the Foreground-Aware Mask (FGM) during the denoising process. In the video generation phase, we propose a two-stage optimization method. In the first stage, we integrate temporal mechanisms into the transformer blocks to maintain effective temporal consistency~\cite{wu2023tuneavideo,gao2023magicdrive}. In the second stage, we introduce a Reward-Guided Diffusion (RGD) framework with Low-Rank Adaptation (LoRA)~\cite{hu2021lora} to ensure high-level consistency. After the denoising process, the frames are passed through the Inception3D (I3D)~\cite{carreira2018quovadisactionrecognition} model to extract high-level features, which are then used to compute the reward function \( R_{\text{I3D}} \). Dense gradients are propagated back to optimize the model, bridging the gap between pixel-level and high-level semantic understanding and promoting high-quality video generation for autonomous driving.

\subsection{Dual-branch Foreground-Background Modeling}
\begin{figure}[t]
\centering{}
\includegraphics[width=1\linewidth]{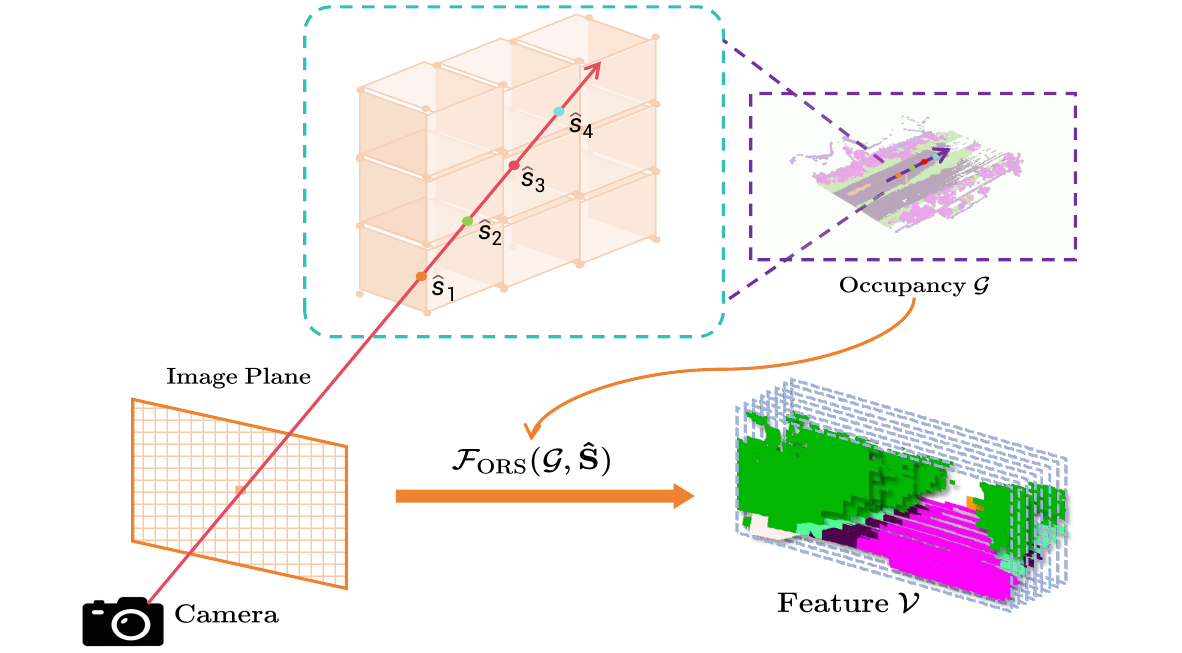}
\caption{Illustrations of our proposed Occupancy Ray-shape Sampling (ORS) method, projecting 3D occupancy grid maps onto the image plane via ray-based sampling, where each pixel is associated with a 3D ray and uniformly sampled points are queried to generate a dense 2D feature representation.} 
\label{fig: ORS}
\vspace{-0.2cm}
\end{figure}

\noindent{\textbf{Occupancy Ray-shape Encoding.}}
Occupancy grid maps are widely utilized as dense 3D representations that encode rich semantic and physical information about the environment. Compared to 3D bounding boxes and binary maps, occupancy grid maps provide models with more precise details and finer-grained control.
 However, considering that the model is a 2D image generation model, directly using the 3D occupancy grid map for control may introduce a domain gap, resulting in suboptimal control performance. To effectively leverage these maps, we propose an Occupancy Ray-shape Sampling (ORS) strategy, which projects the 3D occupancy grid onto the image plane through ray-based sampling, as shown in Fig.~\ref{fig: ORS}. Specifically, for each pixel on the image plane, we associate it with a unique ray \( \boldsymbol{r} \in \mathbb{R}^{3} \) in 3D space, analogous to how optical imaging assigns the color of the first object encountered by a ray to the corresponding pixel in the image plane \cite{whitted2005improved}. In the context of occupancy grid maps, for each ray \( \boldsymbol{r} \), we uniformly sample \( N_{\text{sample}} \) 3D points \( \boldsymbol{\hat{s}}_i \in \mathbb{R}^{3} \) along the ray, spaced by a fixed distance \( n \) (default \( n = 0.2 \) m). The rays and the sampled points are computed as follows:
\begin{equation}
\begin{gathered}
\boldsymbol{r} = \text{Norm}\left( (\boldsymbol{K} \cdot \boldsymbol{T})^{-1} \cdot \boldsymbol{s}_{\text{img}} - \boldsymbol{p}_{\text{ego}} \right) \\
\boldsymbol{\hat{s}}_i = \boldsymbol{p}_{\text{ego}} + \boldsymbol{r} \cdot n \cdot i, \quad i \in \{0, 1, \dots, N_{\text{sample}} \}
\end{gathered}
\end{equation}
where \( \boldsymbol{K} \in \mathbb{R}^{3 \times 3} \) and \( \boldsymbol{T} \in \mathbb{R}^{3 \times 1} \) represent the camera's intrinsic and extrinsic matrices, respectively. \( \text{Norm}(\cdot) \) normalizes the ray vector. The image coordinates \( \boldsymbol{s}_{\text{img}} \in \mathbb{R}^{3} \) are expressed in homogeneous coordinates as \( (u, v, 1) \), where \( u \) and \( v \) represent the pixel coordinates. \( \boldsymbol{p}_{\text{ego}} \in \mathbb{R}^{3} \) denotes the position of the ego vehicle in the occupancy coordinate system. For each sampled point \( \boldsymbol{\hat{s}}_i \), we perform grid sampling by querying the corresponding voxel value from the occupancy grid map \( \mathcal{G} \in \mathbb{R}^{H \times W \times D} \) at the 3D coordinates \( (p_x, p_y, p_z) \) of \( \boldsymbol{\hat{s}}_i \). The ORS strategy treats each sampled point \( \boldsymbol{\hat{s}}_i \) as an index and extracts the corresponding value from the voxel grid \( \mathcal{G} \). This approach allows the sampled points along each ray \( \boldsymbol{r} \) to effectively represent the projection of the voxel grid \( \mathcal{G} \) onto the image plane, facilitating high-quality 3D feature extraction. After processing, a dense feature \( \mathcal{V} \in \mathbb{R}^{U \times V \times N_{\text{sample}}} \) is obtained, which can be expressed as:
\begin{equation}
\mathcal{V} = \mathcal{F}_{\text{ORS}}(\mathcal{G}, \boldsymbol{\hat{S}}),
\end{equation}
 where \( \boldsymbol{\hat{S}} \) denotes the set of all pixels \( \boldsymbol{\hat{s}}_i \) in the image, and \( \mathcal{F}_{\text{ORS}}(\cdot) \) denotes the ORS function, which projects the original voxel grid onto the image plane via ray sampling to produce the corresponding feature representation. In the foreground branch, ORS maps the occupancy grid of the foreground category, \( \mathcal{G}_f \), to the feature \( \mathcal{V}_f = \mathcal{F}_{\text{ORS}}(\mathcal{G}_f, \boldsymbol{\hat{S}}) \). Similarly, in the background branch, ORS maps the occupancy grid of the background category, \( \mathcal{G}_b \), to the feature \( \mathcal{V}_b = \mathcal{F}_{\text{ORS}}(\mathcal{G}_b, \boldsymbol{\hat{S}}) \).

\noindent{\textbf{Driving Scene Encoding.}}
To represent the foreground object \(\mathbf{B}=\{(t^b_i, b_i)\}_{i=1}^{N}\), we utilize 3D bounding box coordinates \(b_i=\{(x_j, y_j, z_j)\}_{j=1}^{8} \in \mathbb{R}^{8 \times 3}\) and the corresponding object category \(t^b \in \mathcal{T}_\text{box}\). For background elements, a vectorized map \(\mathbf{H} = \{(t^m_i, m_i)\}_{i=1}^{N}\) is employed, where \(m_i = \{(v_j)\}_{j=1}^{8} \in \mathbb{R}^{8 \times 3}\) denotes an ordered set of map points (such as street boundaries or crosswalks), and \(t^m \in \mathcal{T}_\text{map}\) represents the category of each map element. To encode the category information, we leverage the CLIP text encoder \cite{radford2021clip}, while the 3D coordinate is processed through Fourier embedding \cite{mildenhall2021nerf}. The final features for both the bounding box and map are obtained by concatenating their respective encodings:
\begin{gather}
\quad \boldsymbol{c}_\text{box} = E_{\text{box}}([\mathrm{CLIP}(t^b), \mathrm{Fourier}(b)]) \\ 
\quad \boldsymbol{c}_\text{map} = E_{\text{map}}([\mathrm{CLIP}(t^m), \mathrm{Fourier}(m)]) 
\end{gather}
For view-specific generation, we incorporate the camera pose \(\mathbf{P} = [\mathbf{K}\in \mathbb{R}^{3 \times 3}, \mathbf{R}\in \mathbb{R}^{3 \times 3}, \mathbf{T}\in \mathbb{R}^{3 \times 1}]\) (intrinsic parameters, rotation, and translation) and a textual sequence \(\mathbf{L}\) to control the abstract semantic content. The features from the text sequence are extracted using a frozen CLIP text encoder, while the camera pose is processed through Fourier embedding. The detailed feature extraction procedure is as follows:
\begin{gather}
\boldsymbol{c}_{\text{text}} = E_{\text{text}}(\mathrm{CLIP}(\mathbf{L})) \\
\boldsymbol{c}_{\text{cam}} = E_{\text{cam}}(\mathrm{Fourier}([\mathbf{K}, \mathbf{R}, \mathbf{T}]^T))
\end{gather}
This approach integrates both geometric and semantic information, enhancing the model's capability for 3D scene generation by providing detailed foreground and background representations, along with context-specific semantic control.

\noindent{\textbf{Foreground Enhancement in DualDiff.}}  
In contrast to previous denoising loss functions used in diffusion models, we propose the Foreground-Aware Mask (FGM) loss, which enhances the model's ability to generate fine-grained obstacle objects, such as distant or intricate structures. FGM dynamically adjusts the weight of the loss function based on the size of the foreground object in the image plane, extending the traditional mean squared error (MSE)~\cite{goodfellow2016deep} loss used in stable diffusion models. Specifically, we construct a mask $\textbf{M}$ using the camera-projected bounding box of the foreground object. The mask assigns higher values to bounding boxes, as defined in the following equation:
\begin{equation}
    {{m}_{ij}} = \left\{
    \begin{array}{ll}
    1.0+\lambda_{fg} - \lambda_{fg}\frac{a_{ij}}{U \times V} & (i,j) \in \text{foreground}, \\
    1.0 & (i,j) \in \text{background}
    \end{array}
    \right.
\end{equation}
where $a_{ij}$ denotes the area of the foreground mask at coordinate $(i,j)$, $\lambda_{fg}$ indicates the weight of attention to the foreground (default 1.0), and $U$ and $V$ represent the width and height of the noisy image feature map, respectively. Finally, to master the denoising process, the network is optimized to predict noise by minimizing the Foreground-Aware Mask error:
\begin{gather}
\begin{split}
\min _\theta \mathcal{L_{FGM}} = &\mathbb{E}_{\boldsymbol{x_0}, \boldsymbol{c}, \tau_\theta(\mathcal{V}_b), \mu_\theta(\mathcal{V}_f), \boldsymbol{\epsilon} \sim \mathcal{N}(0,1), t} \\ 
&\left[ \|\boldsymbol{\epsilon} - \boldsymbol{\epsilon}_{\theta}(z_t, t, \boldsymbol{c}, \boldsymbol{\tau}_\theta(\mathcal{V}_b), \boldsymbol{\mu}_\theta(\mathcal{V}_f))\|_{2}^{2} \right] \odot \mathbf{M}
\end{split}
\end{gather}
where \( \boldsymbol{\epsilon}_{\theta} \) represents the trainable network with parameters \( \theta \), and \( \boldsymbol{c} \) is an optional condition used for conditional generation in DualDiff. This condition consists of two components: the foreground condition \( \boldsymbol{c}_{\text{fg}} \) and the background condition \( \boldsymbol{c}_{\text{bg}} \). Specifically, in the foreground branch \( \boldsymbol{\mu} \), the condition \( \boldsymbol{c}_{\text{fg}} = [\boldsymbol{c}_{\text{cam}}, \boldsymbol{c}_{\text{text}}, \boldsymbol{c}_{\text{box}}] \), and in the background branch \( \boldsymbol{\tau} \), the condition \( \boldsymbol{c}_{\text{bg}} = [\boldsymbol{c}_{\text{cam}}, \boldsymbol{c}_{\text{text}}, \boldsymbol{c}_{\text{map}}] \). The variable $t \in [0, T]$ represents the time step, and $\boldsymbol{\epsilon}$ is additive Gaussian noise. Utilizing the VQ-VAE\cite{esser2020vqvae} encoder, defined as $z = \mathcal{E}(x)$, the feature $z_0$ is diffused over $t$ time steps to obtain the noisy latent $z_t = \sqrt{\bar{\alpha}_t}\boldsymbol{x_0} + \sqrt{1 - \bar{\alpha}_t} \boldsymbol{\epsilon}$, where $\bar{\alpha}_t$ is a scalar parameter. The functions $\boldsymbol{\mu}_\theta(\cdot)$ and $\boldsymbol{\tau}_\theta(\cdot)$ represent the trainable components of the dual-branch foreground-background architecture, while the network $\boldsymbol{\epsilon}_{\theta}$ remains frozen.
\subsection{Multimodal Aware Representation Alignment}
\begin{figure}[t]
\centering{}
\includegraphics[width=1.0\linewidth]{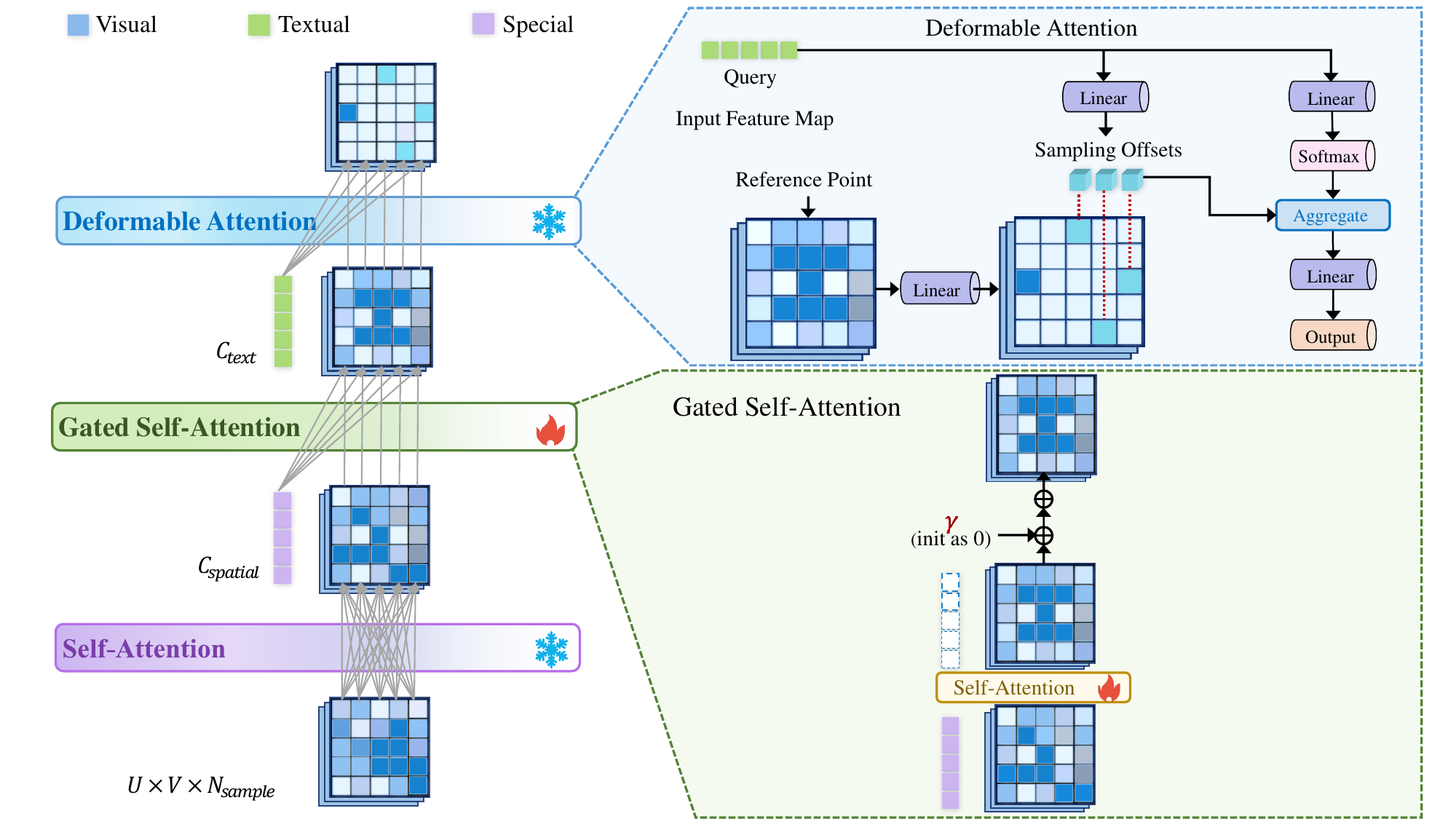}
\caption{Illustration of our proposed Semantic Fusion Attention (SFA) mechanism, which systematically integrates ORS features with Multimodal information. The SFA operates in a sequential manner, enhancing feature representation by leveraging complementary data from various modalities.}
\label{fig: fusion}
\vspace{-0.2cm}
\end{figure}
To align information from different modalities, we propose a Semantic Fusion Attention (SFA) algorithm, which dynamically integrates multimodal information by selectively focusing on relevant features and filtering out noise, thereby enhancing the robustness and accuracy of information fusion. The algorithm combines three types of features: 3D spatial information from $\boldsymbol{c}_\text{box}$ and $\boldsymbol{c}_\text{map}$, rich textual semantics from $\boldsymbol{c}_\text{text}$, and dense visual features represented by $\mathcal{V} \in \mathbb{R}^{U \times V \times N_{\text{sample}}}$, which are extracted through the ORS module. As illustrated in Fig.~\ref{fig: fusion}, we begin by applying a self-attention mechanism~\cite{vaswani2017attention} to the visual features $\mathcal{V}$ obtained from the ORS module, yielding an enhanced visual representation $\mathcal{V}^{\prime}_1 \in \mathbb{R}^{U \times V \times N_{\text{sample}}}$. Next, we concatenate these visual features with spatial features $\boldsymbol{c}_\text{spatial}$, which include both foreground elements $[\boldsymbol{c}_\text{box}, \boldsymbol{c}_\text{cam}]$ and background elements $[\boldsymbol{c}_\text{map}, \boldsymbol{c}_\text{cam}]$. This combined feature set undergoes further processing through a gated self-attention mechanism~\cite{dhingra2017gated}. The design of the gated self-attention mechanism enables dynamic adjustment of attention weights based on the relative importance of the input data, effectively filtering out noise and enhancing spatial localization. Specifically, the gating function \(\tanh(\gamma)\) controls the contribution of each input feature, allowing the model to suppress irrelevant information and maintain accurate spatial information, which results in a refined visual feature $\mathcal{V}^{\prime}_2 \in \mathbb{R}^{U \times V \times N_{\text{sample}}}$. Finally, the spatially enhanced visual features $\mathcal{V}^{\prime}_2$ are fused with textual features through a deformable attention mechanism~\cite{zhu2020deformable}, which adeptly adapts its focus to the complex and dynamic relationships between visual features and textual descriptions, thereby accurately capturing potential spatial correlations. The process culminates in the final output $\mathcal{V}^* \in \mathbb{R}^{U \times V \times N_{\text{sample}}}$, and the entire process is formally described as follows:
\begin{align}
    \mathcal{V}^{\prime}_1 &= \mathcal{V} + \text{SelfAttn}(\mathcal{V}) \\
    \mathcal{V}^{\prime}_2 &= \mathcal{V}^{\prime}_1 + \tanh(\gamma) \cdot \text{SelfAttn}\bigl([\mathcal{V}^{\prime}_1, \boldsymbol{c}_\text{spatial}]\bigr) \\
    \mathcal{V}^* &= \text{DeformAttn}\bigl(\mathcal{V}^{\prime}_2, \boldsymbol{c}_{\text{text}}\bigr)
\end{align}
where $\gamma$ is a learnable scalar (initialized to 0). By integrating multiple modalities such as vision, space, and text, the model effectively captures the complex semantics and dynamics of autonomous driving scenes, producing more realistic, contextually accurate, and geometrically consistent outputs. The gated self-attention design plays a pivotal role in ensuring that the model can efficiently filter out noise and focus on relevant spatial features, ultimately improving the overall performance of multimodal fusion in challenging real-world scenarios.

\subsection{Reward-Guided Diffusion for Video Generation}\label{method:reward-guided}
\begin{figure}[t]
\centering{}
\includegraphics[width=\linewidth]{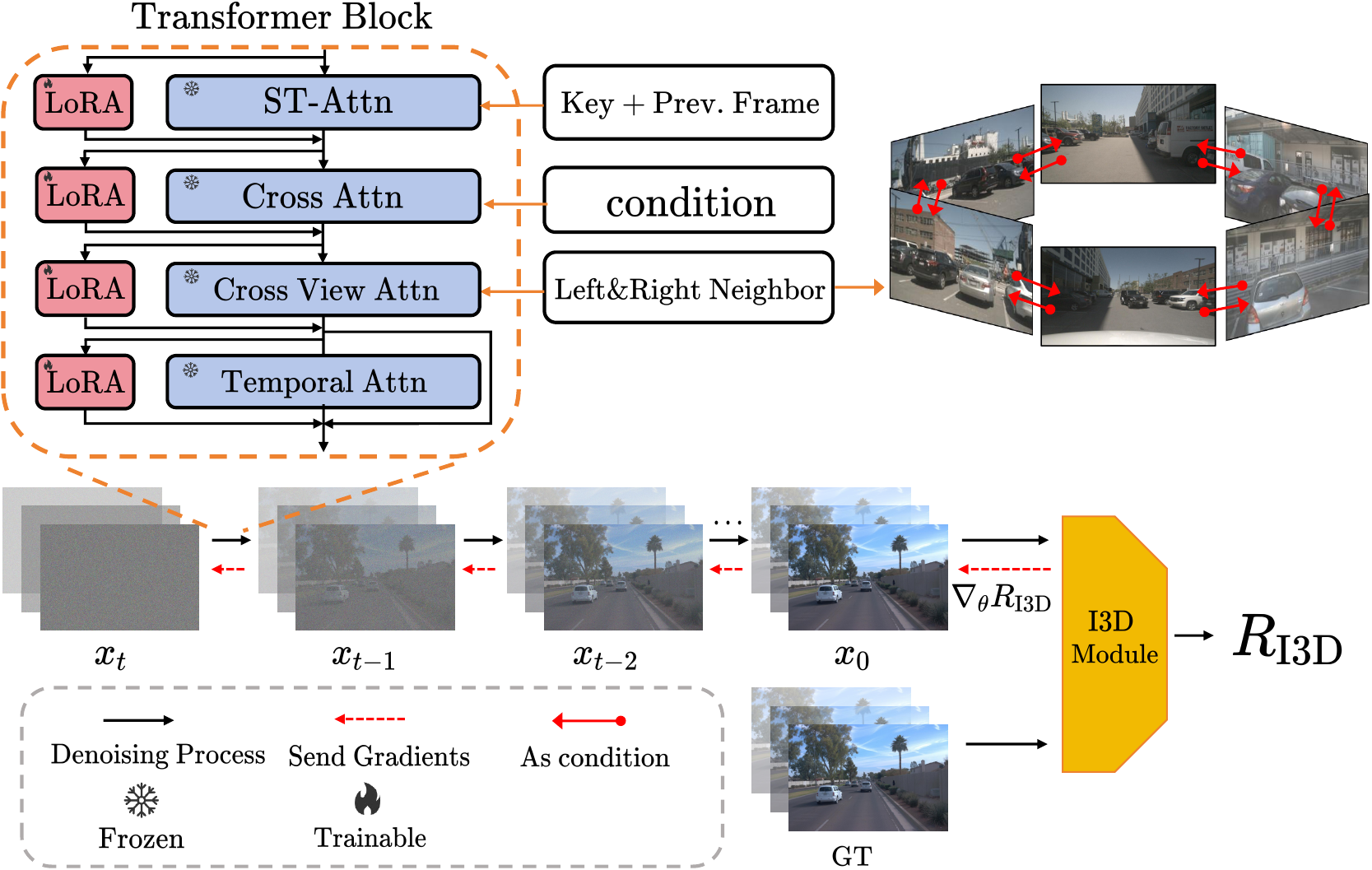}
\caption{\textbf{Overview of the Reward-Guided Diffusion Framework}. For video generation, we extend the panoramic image generation approach by incorporating ST-Attn and Temporal Attn to enhance temporal consistency. In the fine-tuning process, we reduce the number of parameters by adding LoRA to the Attention mechanism of the original network. During training, latent variables are iteratively refined through a denoising loop, starting from pure noise. Denoised frames and ground truth are processed by the I3D model to extract temporal features, which are used to compute the reward function \( R_{\text{I3D}}\). Dense gradients are propagated to optimize the model.}
\label{fig: video}
\vspace{-0.2cm}
\end{figure}
In autonomous driving video generation, existing methods typically rely on pixel-level supervision, often utilizing likelihood loss. While effective at optimizing pixel-wise accuracy, such methods tend to overlook high-level global and structural details that are crucial for generating realistic and contextually accurate videos. To address these issues, we propose a novel Reward-Guided Diffusion (RGD) framework that enhances video generation by incorporating a reward-driven alignment mechanism, as illustrated in Fig.~\ref{fig: video}. The proposed framework utilizes a video diffusion model \( p_\theta(\cdot) \), a video clip dataset \( D_v \), and a reward function \( R(\cdot) \). The reward function is designed to evaluate the generated video based on high-level features, captured through downstream models. Our goal is to maximize the expected reward, defined as:
\begin{equation}
J(\theta) = \mathbb{E}_{c, v \sim D_v, x_0 \sim p_\theta\left( x_0 \mid c \right)} \left[ R\left(x_0, v\right) \right]
\end{equation}
where \( c \) and \( v \) denote the condition and ground truth video clip, respectively, sampled from the dataset \( D_v \), and \( x_0 \) is the generated video conditioned on \( c \). To enable reward-driven optimization, training begins with pure noise \( x_T \), and iterative denoising is carried out following the standard inference procedure of diffusion models. Afterward, gradients are propagated from a differentiable reward function back to the model parameters, enabling the model to learn high-level features. To prevent catastrophic forgetting and preserve the generalization ability of the model, we integrate Low-Rank Adaptation (LoRA)~\cite{hu2021lora} into the attention mechanism of each U-Net layer~\cite{ronneberger2015u}, ensuring stability during the optimization process. The gradient of this process is computed as follows:
\begin{equation}
\nabla_\theta R\left(x_0, v\right) = \sum_{t=0}^T \frac{\partial R\left(x_0, v\right)}{\partial f_t} \frac{\partial f_t}{\partial \theta}
\end{equation}
where \( \theta \) represents the model parameters, and \( f_t \) is the denoising function \( x_{t-1} = f_t(x_t, \theta) \), determined by the denoising scheduler. Upon completing the denoising process, the fully denoised video \( x_0 \) is compared against the ground truth \( v \) using the reward function, which measures the alignment between the generated and reference videos. In our method, the reward function is designed to align high-level semantic features between the generated video and the ground truth. For this purpose, we employ the Inception3D (I3D) model~\cite{carreira2018quovadisactionrecognition} to extract temporal features from both \( x_0 \) and \( v \). The specific reward function $R_{\text{I3D}}$ is defined as the negative distance between their feature representations:
\begin{equation}
R_{\text{I3D}}\left(x_0, v\right) = -\left\| M_{\text{I3D}}\left(x_0\right) - M_{\text{I3D}}\left(v\right) \right\|
\end{equation}
{where \( M_{\text{I3D}}(\cdot) \) is the I3D model designed for extracting temporal features from video inputs. By minimizing the distance between feature representations, the RGD framework ensures the generated videos are not only visually coherent but also semantically aligned with the ground truth. This novel approach effectively bridges the gap between pixel-level supervision and high-level semantic understanding in autonomous driving video generation.}

%% file: sec/4_exp.tex
\section{Experiments}
\label{experiments}
\subsection{Datasets and Baselines}
\noindent\textbf{NuScenes Dataset.}
The nuScenes Dataset \cite{caesar2020nuscenes}, developed by Aptiv, is a large-scale multimodal dataset for the research of autonomous driving. Training data is collected by 1 spinning LIDAR, 5 long-range RADAR sensors, and 6 cameras. The nuScenes dataset includes 1000 scenes, which are then split into 700/150/150 scenes for training/validation/testing.

\noindent{\textbf{Waymo Dataset.}}
The Waymo Dataset~\cite{Sun_2020_waymo} is a large-scale, multimodal dataset created for autonomous driving research, containing over 1,000 hours of driving data collected in diverse urban and suburban environments. It includes sensor data from high-definition (HD) cameras, LiDAR, and RADAR, which provide rich input for a variety of autonomous driving tasks. The dataset is annotated with detailed information such as vehicle trajectories, 3D object detection labels, lane markings, and traffic signal data, making it suitable for tasks like object detection, tracking, scene segmentation, and behavior prediction.

\noindent{\textbf{Baselines.}}
Our baseline builds on recent advancements in generating street-view images and videos for autonomous driving, integrating several classic state-of-the-art (SOTA) methods. For instance, MagicDrive~\cite{gao2023magicdrive} emphasizes end-to-end autonomous driving with enhanced spatiotemporal perception, Panacea~\cite{wen2024panacea} introduces a unified framework for multi-task learning in driving scenarios, and Drive-WM~\cite{wang2024drivewm} demonstrates robust performance under challenging weather and motion uncertainties through advanced sensor fusion techniques.

\subsection{Evaluation Metrics.}
To evaluate the quality of our synthesized data, we use frame-wise Fr{\'e}chet Inception Distance (FID)~\cite{heusel2017gans} and Fr{\'e}chet Video Distance (FVD)~\cite{unterthiner2018fvd}. FID assesses the fidelity of individual frames, while FVD captures both image quality and temporal consistency. The controllability of DualDiff is demonstrated by the alignment between generated videos and conditioned BEV sequences. We measure foreground feature acquisition on the nuScenes~\cite{caesar2020nuscenes} and Waymo~\cite{Sun_2020_waymo} datasets using metrics such as nuScenes Detection Score (NDS), mean Average Precision (mAP), mean Average Orientation Error (mAOE), and mean Average Velocity Error (mAVE). Background features, including road and vehicle segmentation, are evaluated using mean Intersection over Union (mIoU). Our evaluation framework consists of two components: (1) comparing the performance of synthesized data with real-world data using a pre-trained perception model, and (2) exploring the potential of synthesized data as an augmentation strategy to enhance training. This dual evaluation provides insights into the quality and utility of the generated data.

For image-based generation, we tested CVT~\cite{zhou2022cvt} and BEVFusion~\cite{liu2022bevfusion} on the nuScenes dataset with pre-trained models. For the Waymo dataset, we used BEVFormer~\cite{li2022bevformer}. For video-based methods, we used BEVFormer~\cite{li2022bevformer}, and the final score~\cite{coda2024} was computed as:
\begin{equation}
\text{score} = \frac{a - \mathrm{FVD}}{a} + \frac{\mathrm{mAP} - b}{b} + \frac{\mathrm{mIoU} - c}{c}
\end{equation}
where \( a = 218.1200 \), \( b = 11.8617 \), and \( c = 18.3429 \). Finally, we evaluated the model training performance using StreamPETR~\cite{wang2023streampetr}, a state-of-the-art video-based perception method, for the data-centric closed loop.

\subsection{Implementation Details}
We implement our approach using Stable Diffusion v1.5~\cite{rombach2022high} with ControlNet weights for segmentation tasks, keeping the U-Net~\cite{ronneberger2015u} frozen throughout training. The model is trained on 8 A800 GPUs, initially training the dual foreground and background branches separately for 90k steps, followed by fine-tuning for 30k additional steps. 1) For image sampling, we use UniPC~\cite{zhao2024unipc} with 20 sampling steps and a Classifier-Free Guidance (CFG)~\cite{ho2022classifier} of 2.0. The image resolution is 224 $\times$ 400 for the nuScenes~\cite{caesar2020nuscenes} dataset and 320 $\times$ 480 for Waymo~\cite{Sun_2020_waymo}. 2) For video generation, we generate 16-frame videos using the 12Hz nuScenes~\cite{caesar2020nuscenes} dataset. The ControlNet~\cite{zhang2023controlnet} is frozen, while the Spatio-Temporal Attention (ST-Attn)~\cite{wu2023tuneavideo} and Temporal Attention (Temporal Attn)~\cite{gao2023magicdrive} components of the U-Net~\cite{ronneberger2015u} are trained. 3) To further enhance the model, we apply Low-Rank Adaptation (LoRA)~\cite{hu2021lora} to all attention layers for reward model fine-tuning, while keeping the original parameters fixed. Gradient checkpointing manages GPU memory, and UniPC~\cite{zhao2024unipc} is configured with 10 sampling steps during fine-tuning, ensuring efficient resource utilization and high-quality output for both images and videos.

\subsection{Main Results}
\noindent{\textbf{Comparison with Baselines.}}
We conducted a comparative analysis of several baseline methods on the nuScenes~\cite{caesar2020nuscenes} dataset. Under consistent resolution settings, DuaDiff outperformed MagicDrive~\cite{gao2023magicdrive} in reconstructing realistic street scene styles, achieving a 5.6\% reduction in FID at 224×400 resolution and 3.42\% at higher resolutions (Table~\ref{tab:main}). Table~\ref{tab:fvd} further highlights the superior performance of DuaDiff. Evaluating on the Waymo~\cite{Sun_2020_waymo} dataset, DuaDiff showed even greater improvements, reducing FID by 5.71\% at 224×400 resolution (Table~\ref{tab:main_waymo}). For 3D object detection in the 0–30m range, it improved mAP by 7.3\%, while for 3D segmentation, DuaDiff achieved gains of 0.79\% and 0.62\% in mIoU for the Road and Vehicle categories, respectively. These results demonstrate the robustness and generalization capabilities of DuaDiff across different datasets and tasks.

\noindent{\textbf{Parameter Quantity Evaluation.}}\label{ex:main_pq}
To isolate the impact of the dual-branch design, ablation studies (Table~\ref{tab:ablation}) compare DuaDiff with a variant where Semantic Fusion Attention (SFA) outputs are summed and fed into ControlNet~\cite{zhang2023controlnet} (w/n decouple). With comparable parameters, DuaDiff achieves a 1.23\% FID reduction, a 1.08\% mAP increase in 3D detection, and a 0.84\% mIoU improvement in BEV segmentation, validating the intrinsic advantages of its decoupled dual-branch. 


\begin{table*}
\caption{Comparison of generation fidelity among various driving-view generation methods. The synthesis conditions are derived from the nuScenes validation set, and each task employs models trained on the corresponding nuScenes training set. DualDiff consistently outperforms all baseline models across the evaluation metrics. The best results are in \textbf{bold}, while the second best results are in \uline{underlined italic}.} \label{tab:main}
\centering{}%
\scalebox{1.0}{
\begin{tabular}{l|c|c|c|c|c|c|c}
\toprule
  \multirow{2}[3]{*}{Methods} &
  \multirow{2}[3]{*}{\begin{tabular}[c]{@{}c@{}}Synthesis\\ resolution\end{tabular}} &
  \multirow{2}[3]{*}{FID$ \downarrow$} &
  \multicolumn{2}{c|}{BEV Segmentation} &
  \multicolumn{2}{c|}{3D Object Detection} & 
  \multirow{2}[3]{*}{Avene}\\
  \cmidrule{4-7} 
  &&&
  Road mIoU $\uparrow$ &
  Vehicle mIoU $\uparrow$ &
  mAP $\uparrow$ &
  NDS $\uparrow$   \\
\midrule
\textcolor{gray}{Oracle\cite{gao2023magicdrive}}
    & \textcolor{gray}{-}
    & \textcolor{gray}{-}
    & \textcolor{gray}{72.21}
    & \textcolor{gray}{33.66}
    & \textcolor{gray}{35.54}
    & \textcolor{gray}{41.21} 
    & \textcolor{gray}{ICLR2024}\\
\textcolor{gray}{Oracle\cite{gao2023magicdrive}}
    & \textcolor{gray}{224$\times$400}
    & \textcolor{gray}{-}
    & \textcolor{gray}{72.19}
    & \textcolor{gray}{33.61}
    & \textcolor{gray}{23.54}
    & \textcolor{gray}{31.08} 
    & \textcolor{gray}{ICLR2024}\\
\midrule
BEVGen\cite{swerdlow2024bevgen}
    & 224$\times$400
    & 25.54
    & 50.20
    & 5.89
    & -
    & -     
    & RAL\textcolor{blue}{2024} \\
BEVControl\cite{yang2023bevcontrol}
    & -
    & 24.85
    & 60.80
    & 26.80
    & -
    & -    
    & Arxiv\textcolor{blue}{2023} \\ 
MagicDrive\cite{gao2023magicdrive}
    & 224$\times$400
    & 16.20
    & 61.05
    & 27.01
    & 12.30
    & 23.32 
    & ICLR\textcolor{blue}{2024} \\
MagicDrive\cite{gao2023magicdrive}
    & 272$\times$736
    & 16.59
    & 54.24
    & \uline{31.05} \textcolor{darkgreen}{(-2.6\%)}
    & \uline{20.85} \textcolor{darkgreen}{(-2.7\%)}
    & \uline{30.26} \textcolor{darkgreen}{(-0.8\%)}
    & ICLR\textcolor{blue}{2024} \\
PerlDiff\cite{zhang2024perldiff}
    & 256$\times$704
    & 25.06
    & 61.26
    & 27.13
    & 15.24
    & 24.05 
    & Arxiv\textcolor{blue}{2024} \\
\midrule
\textbf{DualDiff} (\textbf{Ours})
    & 224$\times$400
    & \textbf{10.99}
    & \textbf{62.75} \textcolor{darkgreen}{(-9.4\%)}
    & 30.22
    & 13.99
    & 24.98
    & \textcolor{blue}{2024} \\
\textbf{DualDiff} (\textbf{Ours})
    & 432$\times$768
    & \uline{13.16}
    & \uline{62.38} \textcolor{darkgreen}{(-9.8\%)}
    & \textbf{31.69} \textcolor{darkgreen}{(-1.9\%)}
    & \textbf{22.13} \textcolor{darkgreen}{(-1.4\%)}
    & \textbf{30.96} \textcolor{darkgreen}{(-0.1\%)}
    & \textcolor{blue}{2024} \\
\bottomrule
\end{tabular}
}
\vspace{-0.2cm}
\end{table*}

\begin{table*}[t]
\caption{Comprehensive comparison of generation fidelity across previous methods trained with nuScenes. DualDiff outperforms the state-of-the-art (SOTA) street scene reconstruction models on nuScenes validation set. The best results are in \textbf{bold}, while the second best results are in \uline{underlined italic}.}
\label{tab:fvd}
\vspace{-3mm}
\scalebox{0.75}{
\begin{tabular}[t]{c|cccccccccc|c}
\toprule
\multirow{2}{*}{Metrics} 
& DriveGAN\cite{kim2021drivegan}
& GenAD \cite{zheng2025genad} 
& PerlDiff\cite{zhang2024perldiff}
& DriveDreamer-2\cite{zhao2024drivedreamer2}
& DrivingDiffuion\cite{li2025drivingdiffusion}
& Panacea+\cite{wen2024panacea} 
& MagicDrive\cite{gao2023magicdrive} 
& SimGen\cite{zhou2024simgen}
& Delphi\cite{ma2024delphi}
& Drive-WM\cite{wang2024drivewm}
& \cellcolor{gray!20}\textbf{DualDiff}\\
& CVPR\textcolor{blue}{2021}
& CVPR\textcolor{blue}{2024} 
& Arxiv\textcolor{blue}{2024}
& Arxiv\textcolor{blue}{2024}
& ECCV\textcolor{blue}{2023}
& CVPR\textcolor{blue}{2024}
& ICLR\textcolor{blue}{2024} 
& Arxiv\textcolor{blue}{2024}
& Arxiv\textcolor{blue}{2024}
& CVPR\textcolor{blue}{2024}
& \cellcolor{gray!20}(\textbf{Ours}) \\
\midrule
\midrule
FID$\downarrow$ 
& 73.4 
& 15.4 
& 25.06 
& 25.0 
& 15.8 
& 15.5 
& 16.20 
& 15.60 
& \uline{15.08} 
& 15.80 
& \cellcolor{gray!20}\textbf{10.99}\\
FVD$\downarrow$ 
& 502 
& 244 
& - 
& \uline{105}
& 332
& \textbf{103}
& 237
& 271
& 114
& 123
& \cellcolor{gray!20}160\\
\bottomrule
\end{tabular}
}
\vspace{-6mm}
\label{tab:fid}
\end{table*}


\begin{table}
\caption{Performance comparison on the Waymo validation set. Our proposed DualDiff algorithm significantly outperforms the mainstream MagicDrive algorithm in terms of generation fidelity and performance on downstream foreground and background tasks. The metric ``$\text{mAP}_{0\sim30}$" represents the average Average Precision (AP) for vehicles, pedestrians, and cyclists within $0\sim30$ meters of the ego. The best results are highlighted in \textbf{bold}.} 
\label{tab:main_waymo}
\centering{}%
\scalebox{0.9}{
\begin{tabular}{l|c|c|c|c}
\toprule
\multirow{2}[1]{*}{Methods} & \multirow{2}[1]{*}{FID$\downarrow$} & \multicolumn{2}{c|}{BEV Segmentation} & \multicolumn{1}{c}{3D Detection} \\
\cmidrule{3-5}
 & & Road mIoU $\uparrow$ & Vehicle mIoU $\uparrow$ & $\text{AP}_{0\sim30}$ $\uparrow$ \\
\midrule
\midrule
\textcolor{gray}{Oracle\cite{gao2023magicdrive}}
    & \textcolor{gray}{-}
    & \textcolor{gray}{61.20}
    & \textcolor{gray}{53.65}
    & \textcolor{gray}{24.9} \\
\midrule
MagicDrive\cite{gao2023magicdrive}
    & 17.16 
    & 50.45
    & 47.93
    & 3.1 \\
\textbf{DualDiff} (\textbf{Ours})
    & \textbf{11.45}
    & \textbf{51.24} 
    & \textbf{48.55} 
    & \textbf{10.4} \\
\bottomrule
\end{tabular}
}
\vspace{-0.4cm}
\end{table}

\noindent{\textbf{Training Support for Downstream Perception Tasks.}}\label{ex:main_more_downstream}
To comprehensively assess the quality of the generated images, we conducted an evaluation on downstream perception tasks, as shown in Table~\ref{tab:downstream}. DuaDiff was used to generate an evaluation set of the same size as the original real-world dataset, and the CVT~\cite{zhou2022cvt} and BEVFusion~\cite{liu2022bevfusion} models, trained on real data, were employed to assess foreground and background performance. In the 3D object detection task, DuaDiff improved mAP by 1.46\% compared to Drive-WM~\cite{wang2024drivewm}. For BEV segmentation, DuaDiff outperformed Drive-WM by 4.50\% in Vehicle mIoU. Compared to MagicDrive~\cite{gao2023magicdrive}, it improved Vehicle mIoU by 4.68\% and Road mIoU by 1.70\%. These results highlight the robustness and versatility of DuaDiff across various tasks, demonstrating its effectiveness in generating high-quality synthetic data that enhances model performance for real-world applications. The significant improvements across these tasks further affirm that DuaDiff provides a valuable tool for advancing perception systems in complex environments.

\begin{table}
\caption{Evaluation of DuaDiff on downstream perception tasks, including 3D object detection and background segmentation, showing improvements in mAP and mIoU compared to baseline methods. The best results are in \textbf{bold}, and the second-best results are in \uline{underlined italic}.}
\centering{}%
\scalebox{0.90}{
\begin{tabular}{l|c|c|c}
    \toprule
     \multirow{2}[1]{*}{Methods} & 
     \multicolumn{1}{c|}{3D Detection} & 
     \multicolumn{2}{c}{BEV Segmentation} \\
     \cmidrule{2-4}
    &
    mAP $\uparrow$ &
    Vehicle mIoU $\uparrow$ &
    Road mIoU $\uparrow$ \\
    \midrule
    \midrule
    \textcolor{gray} {Oracle~\cite{wang2024drivewm}} & \textcolor{gray}{37.78} &  \textcolor{gray}{36.08}&\textcolor{gray}{72.36}\\
    \midrule
    BEVGen~\cite{swerdlow2024bevgen} &- &5.89 & 50.20  \\
    LayoutDiffusion~\cite{zheng2023layoutdiffusion} & 3.68 & 15.51 & 35.31 \\
    GLIGEN~\cite{li2023gligen} & 15.42 &22.02 & 38.12 \\ 
    BEVControl~\cite{yang2023bevcontrol} & 19.64 & 26.80 & 60.80 \\
    MagicDrive~\cite{gao2023magicdrive} & 12.30 & 27.01 & \uline{61.05} \textcolor{darkgreen}{(-11.3\%)}\\
    Drive-WM~\cite{wang2024drivewm} & \uline{20.66} \textcolor{darkgreen}{(-17.12\%)} & \uline{27.19} \textcolor{darkgreen}{(-8.9\%)} & - \\
    \midrule
    \cellcolor{gray!20}\textbf{DualDiff} (\textbf{Ours}) & \cellcolor{gray!20}\textbf{22.13} \textcolor{darkgreen}{(-15.6\%)}& \cellcolor{gray!20}\textbf{31.69} \textcolor{darkgreen}{(-4.4\%)}& \cellcolor{gray!20}\textbf{62.75} \textcolor{darkgreen}{(-9.6\%)}\\
    \bottomrule
  \end{tabular}
}
\label{tab:downstream}
\vspace{-0.4cm}
\end{table}

\noindent{\textbf{Quality Evaluation of Generated Videos.}}\label{ex:main_video_quality}
As shown in Table \ref{tab:video_fvd}, our method significantly outperforms the baseline (MagicDrive~\cite{gao2023magicdrive}) on downstream video perception tasks, achieving a 1.46\% improvement in 3D detection and a 3.88\% gain in BEV segmentation. The combined metric of FVD, mAP, and mIoU shows a relative improvement of 0.32\%. Notably, on the NuScenes~\cite{caesar2020nuscenes} validation set, our approach reduces the FVD score by 32.5\% compared to MagicDrive, demonstrating its ability to generate high-fidelity videos with accurate foreground and background details. Additionally, our Reward-Guided Diffusion (RGD) method enhances video generation quality, reducing FVD by 86.06 and improving the overall evaluation score by 0.39 under the same training iterations. These results highlight RGD's impact on temporal consistency, task relevance, and video quality, advancing the state of the art in video synthesis for driving scenarios.

\begin{table}
\caption{Comparations on Video Generation: We evaluate 150 cases from NuScenes evaluation set (aligned with the evaluation protocol of ECCV2024 Workshop \cite{coda2024}), reporting FVD scores and downstream task performance using the BEVFormer~\cite{li2022bevformer}. 
}
\centering{}%
\scalebox{0.83}{
\begin{tabular}{l|c|c|c|c}
\toprule
 \multirow{2}[1]{*}{Methods} & 
 \multirow{2}[1]{*}{FVD$\downarrow$} & 
 \multicolumn{1}{c|}{3D Detection} & 
 \multicolumn{1}{c|}{BEV Segmentation} &
 \multirow{2}[1]{*}{Final Score} \\
 \cmidrule{3-4}
 & & mAP $\uparrow$ & Road mIoU $\uparrow$ & \\
\midrule
\midrule
Seven  \cite{coda2024}& 232.50 & 12.78 & 19.46 & 0.07 \\
MagicDrive \cite{gao2023magicdrive} & \textbf{218.12} & 11.86 & 18.34 & 0 \\
\midrule
DualDiff (w/n reward) & 306.07 & 13.29 & \textbf{22.25} & -0.07 \\
\cellcolor{gray!20}\textbf{DualDiff} (\textbf{w reward}) & \cellcolor{gray!20}220.01 & \cellcolor{gray!20}\textbf{13.32} & \cellcolor{gray!20}22.22 & \cellcolor{gray!20}\textbf{0.32} \\
\bottomrule
\end{tabular}
}
\label{tab:video_fvd}
\vspace{-0.4cm}
\end{table}

\subsection{Data-centric Closed-loop Training and Evaluation}\label{ex:closed_loop}
\begin{figure}
\centering{}
\includegraphics[width=1.0\linewidth]{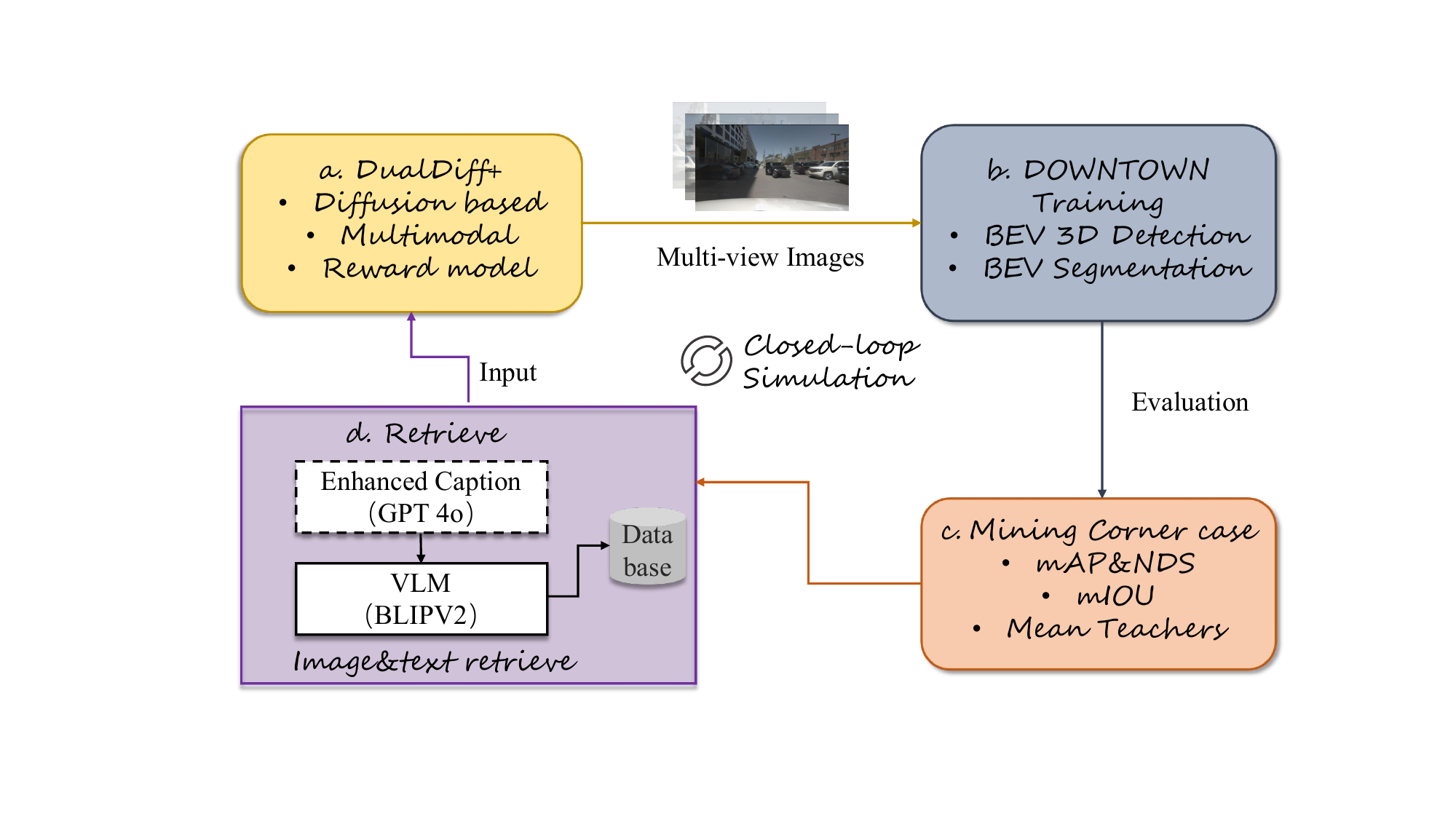}
\caption{A comprehensive data-centric closed-loop framework comprising four key components: a. Image and video generation; b. Downstream task training; c. Corner case mining; d. Hard example retrieval.} 
\label{fig:close-lop}
\vspace{-0.2cm}
\end{figure}

\begin{table*}[t]
\centering
\caption{
Performance comparison of 3D object detection by fine-tuning the StreamPETR~\cite{wang2023streampetr} open-source model using various data sampling strategies, data volumes, generative model techniques, and data sources within a data-centric closed-loop framework. The first row presents the baseline results for reference.
}
\scriptsize
\resizebox{1.0\textwidth}{!}{
\begin{tabular}{c|c|c|c|cccc}
\toprule
\multirow{2}{*}{Sampling Strategy} & \multirow{2}{*}{Num of Cases (k)} & \multirow{2}{*}{Methods} & \multirow{2}{*}{Data Source} & \multicolumn{4}{c}{3D Object Detection} \\
\cmidrule{5-8}
& & & & mAP$\uparrow$ & mAOE$\downarrow$ & mAVE$\downarrow$ & NDS$\uparrow$ \\ 
\midrule
\textcolor{gray}{Baseline (StreamPETR~\cite{wang2023streampetr})} & \textcolor{gray}{--} & \textcolor{gray}{--} & \textcolor{gray}{nuScenes} & \textcolor{gray}{34.5} & \textcolor{gray}{59.4} & \textcolor{gray}{29.1} & \textcolor{gray}{46.9} \\ 
\midrule
\multirow{4}{*}{+ Random Sampling} 
& \cellcolor{gray!10} 2.4 (10\%) & \cellcolor{gray!10} MagicDrive~\cite{gao2023magicdrive} & \cellcolor{gray!10} {nuScenes} & \cellcolor{gray!10}32.90 & \cellcolor{gray!10}53.35 & \cellcolor{gray!10}32.35 & \cellcolor{gray!10}46.30 \\
& 2.4 (10\%) & \multirow{3}{*}{DualDiff} & \multirow{2}{*}{nuScenes} & 34.68 & 53.66 & 31.24 & 47.44 \\
& 4.8 (20\%) & & & 34.71 \textcolor{red}{(+0.03\%)} & 53.22 \textcolor{darkgreen}{(-0.44\%)} & 30.52 \textcolor{darkgreen}{(-0.71\%)} & 47.92 \textcolor{red}{(+0.48\%)} \\ %
& 2.4 (10\%) & & Waymo & 36.66 & - & - & - \\
\midrule
\multirow{4}{*}{\begin{tabular}[c]{@{}l@{}}+ Corner-case \\ Driven Sampling\end{tabular}} 
& \cellcolor{gray!10}2.4 (10\%) & \cellcolor{gray!10}MagicDrive~\cite{gao2023magicdrive} & \cellcolor{gray!10}nuScenes & \cellcolor{gray!10}34.60 & \cellcolor{gray!10}56.19 & \cellcolor{gray!10}29.17 & \cellcolor{gray!10}47.30 \\
& 2.4 (10\%) & \multirow{3}{*}{DualDiff} & \multirow{2}{*}{nuScenes} & 36.74 & 48.09 & 28.80 & 49.50 \\
& 4.8 (20\%) & & & 37.69 \textcolor{red}{(+0.94\%)}  & 45.38 \textcolor{darkgreen}{(-2.71\%)} & 27.19 \textcolor{darkgreen}{(-1.60\%)} & 50.87 \textcolor{red}{(+1.36\%)} \\
& 2.4 (10\%) & & Waymo & 38.00 & - & - & - \\ 
\bottomrule
\end{tabular}
}
\label{tab:failure_case_driven_framework}
\vspace{-0.2cm}
\end{table*}
To effectively leverage generated data, a common approach is to randomly select a subset of the training set and apply a generative model to enhance it, aiming to improve downstream task performance. However, we argue that incorporating corner cases is often more crucial than simply increasing the volume of common data. To address this, we propose a data-centric closed-loop framework (Fig. \ref{fig:close-lop}) that uses a generative model to identify challenging samples for fine-tuning, ultimately boosting performance on downstream perception tasks. The framework consists of four key steps: (1) use the evaluation set to identify failure cases from the current model, (2) apply a visual-language-based method with the multimodal model BLIPv2~\cite{li2023blip2} to analyze patterns and retrieve similar scenes, (3) diversify scene and instance-level captions to generate new data with varied appearances, and (4) fine-tune the model with this new data to enhance generalization.

As shown in Table~\ref{tab:failure_case_driven_framework}, we first trained the model on the full nuScenes dataset to establish baseline performance (first row) for downstream 3D object detection metrics. We then augmented the training set using two data sampling methods. Experimental results show that the corner-case driven sampling method, based on the data-centric closed-loop framework, improves performance by 2\% compared to random sampling. Specifically, detailed ablation experiments reveal: 1) When adding 10\% more nuScenes~\cite{caesar2020nuscenes} data with the same image generation algorithm, the corner-case driven method improves mAP by 1.7\% on MagicDrive~\cite{gao2023magicdrive} and by 2.06\% on DualDiff compared to random sampling. 2) Adding 10\% of Waymo~\cite{Sun_2020_waymo} data yields a 1.34\% improvement in mAP. 3) While random sampling shows limited improvement with larger datasets, the data-centric closed-loop method effectively identifies and learns from challenging samples, resulting in significant performance gains, with mAP and NDS improving by 0.94\% and 1.36\%, respectively. These results highlight the superiority of corner-case driven sampling over random sampling in enhancing model performance.

\vspace{-0.2cm}
\subsection{Ablation Studies}
\begin{figure}
\centering{} 
\includegraphics[width=1.0\columnwidth]{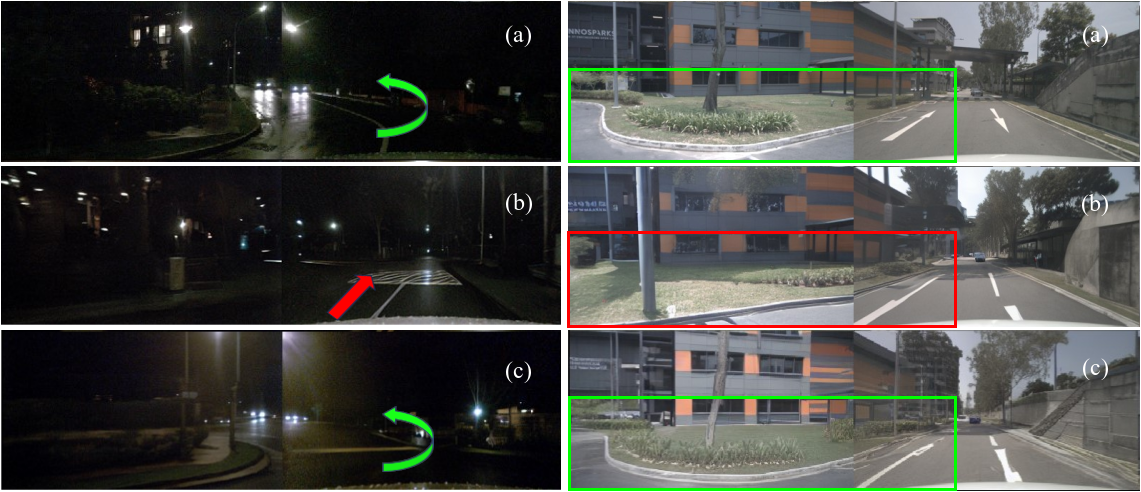}
\caption{Driving scenes of (a) ground truth, (b) MagicDrive, and (c) DuaDiff (Ours). Compared to the baseline, DuaDiff accurately reproduces the left-turn orientation and the car in the distance in the night scene, while in the daylight scene, it precisely generates the road edge and the tree in the background.}
\vspace{-0.4cm}
\label{fig:road_night_case}
\end{figure}

\begin{figure}
\centering{}
\includegraphics[width=0.95\columnwidth]{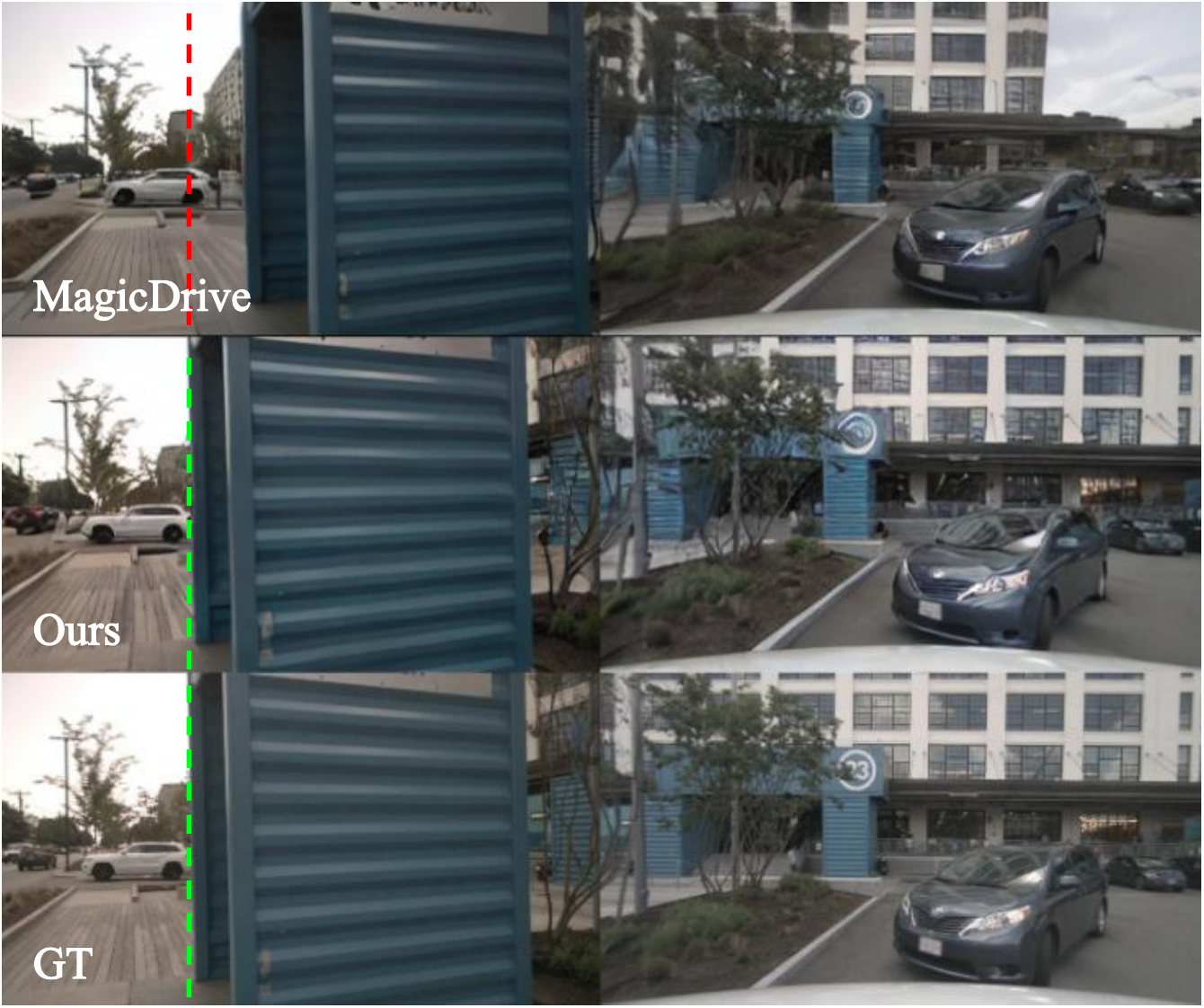}
\caption{Daytime driving scene, where our model accurately generates foreground information through Occupancy Ray-shape Sampling (ORS), improving geometry spatial accuracy.}
\label{fig:ors_case}
\vspace{-0.4cm}
\end{figure}

\noindent{\textbf{Occupancy Ray-shape Encoding.}}
In contrast to the baseline MagicDrive~\cite{gao2023magicdrive}, which relies on BEV road map information, we replace road map encoding with Occupancy Ray-shape Sampling (ORS) features as conditioning inputs. The results presented in the second row of Table~\ref{tab:ablation} demonstrate substantial improvements with the use of ORS features, outperforming the road map-based background encoding. Specifically, the FID score decreases by 2.94\%, while the background road mIoU improves by 1.14\%. As illustrated in Fig.~\ref{fig:road_night_case}, our model generates road layouts with precise edge details. Compared to the BEV road map encoding, ORS features ensure viewpoint consistency with the ground truth street view images, facilitating faster model convergence. Beyond viewpoint alignment, ORS also enhances the accuracy of spatial geometry details. For instance, as shown in Fig.~\ref{fig:ors_case}, our model demonstrates improved control over object positioning, providing more detailed layout information that benefits downstream tasks.

\begin{figure}
\centering{}
\includegraphics[width=1.0\linewidth]{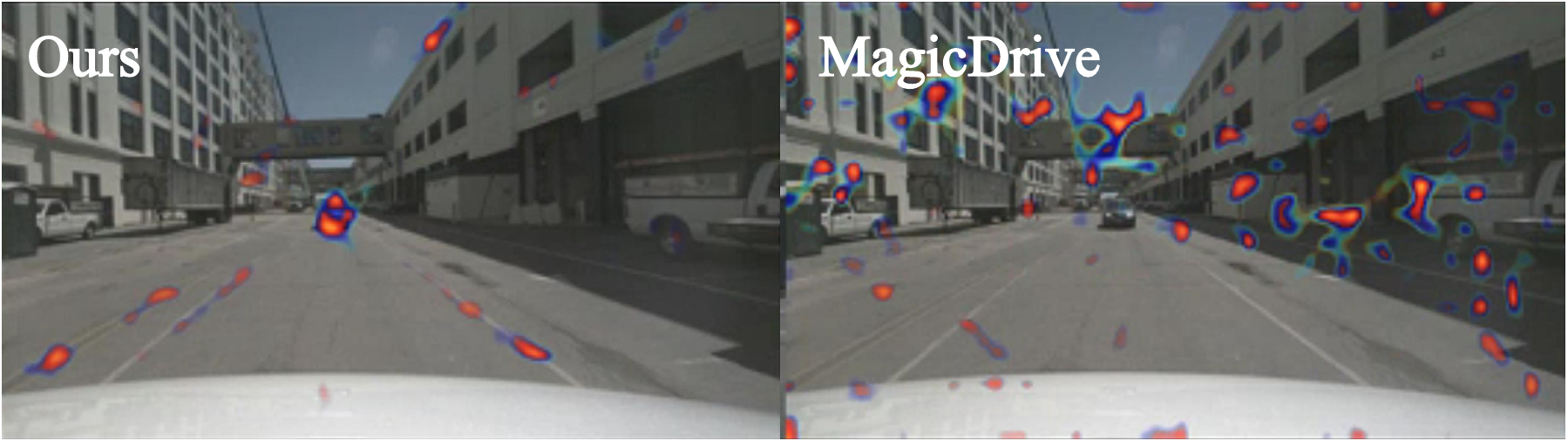}
\caption{Visualization of attention. 
With Semantic Fusion Attention (SFA), our model places greater emphasis on both foreground (e.g., vehicles) and background (e.g., lane markings) features, in contrast to MagicDrive.} 
\label{fig:sfa-vis}
\vspace{-0.4cm}
\end{figure}

\begin{figure}
\centering{}
\includegraphics[width=0.95\columnwidth]{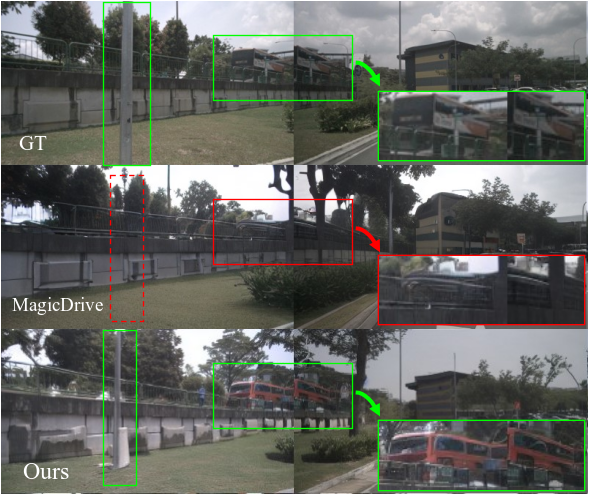}
\caption{Reconstruction of the scene in daylight, where our model accurately generates the bus in the distance and the lamp pole, preserving both the spatial arrangement and fine details.}
\label{fig:day_road_case}
\vspace{-0.4cm}
\end{figure}

\noindent{\textbf{Ablating FGM and SFA.}}
In Table \ref{tab:ablation}, we first evaluate the performance of the Foreground-Aware Mask (FGM) module without employing a dual-branch setup. Experimental results demonstrate that the FGM module significantly improves foreground object detection, with a 0.4\% increase in mAP for downstream 3D detection and a 1.70\% improvement in the mIoU of foreground vehicles in BEV segmentation. Under the dual-branch configuration, the mAP for 3D detection increases by an additional 0.83\%, and the mIoU for foreground vehicles in BEV segmentation improves by 1.02\%. As shown in Fig. \ref{fig:day_road_case}, the FGM module effectively enhances the generation of distant obstacles. Next, we also conduct a detailed analysis of the Semantic Fusion Attention (SFA) module, which leads to a 1.58\% reduction in FID, a 1.67\% improvement in foreground mAP, and a 2.92\% increase in BEV segmentation vehicle mIoU. Fig. \ref{fig:sfa-vis} clearly illustrates the effectiveness of the gated self-attention mechanism within SFA, which dynamically adjusts attention weights based on the relative importance of the input data. This mechanism not only suppresses noise but also enhances spatial positioning information.

\begin{table}
\caption{Ablation study of the proposed module, evaluating its impact on FID scores, downstream 3D detection performance (BEVFusion~\cite{liu2022bevfusion}), and segmentation tasks (CVT~\cite{zhou2022cvt}). 
}
\centering{}%
\scalebox{0.7}{
\begin{tabular}{c|ccc|c|c|c|c}
\toprule
\multirow{2}[1]{*}{Dual-Branch} & 
\multirow{2}[1]{*}{SFA} & 
\multirow{2}[1]{*}{FGM} & 
\multirow{2}[1]{*}{ORS} & 
\multirow{2}[1]{*}{FID $\downarrow$} & 
 \multicolumn{1}{c|}{3D Detection} & 
 \multicolumn{2}{c}{BEV Segmentation} \\
 \cmidrule{6-8}
&&&&&
mAP $\uparrow$ & 
Road mIoU $\uparrow$ & 
Vehicle mIoU $\uparrow$ \\
\midrule
\midrule
& & & & \textcolor{gray}{16.20} & \textcolor{gray}{12.30} & \textcolor{gray}{61.05} & \textcolor{gray}{27.01} \\
& & & \checkmark & 13.26 & 12.32 & 62.19 & 27.30 \\
& & \checkmark & \checkmark & 12.95 & 12.71 & 61.78 & 29.00 \\
&  \checkmark & \checkmark & \checkmark & 12.57 & 12.83 & 61.47 & 29.11 \\
\checkmark & \checkmark & & \checkmark & {11.01} & 13.16 & {62.71} & {29.19} \\
\midrule
\checkmark (w/n decouple) & \checkmark & \checkmark & \checkmark & 12.22 & 12.91 & 62.30 & 29.27 \\
\cellcolor{gray!20}\checkmark (\textbf{Ours}) & \cellcolor{gray!20}\checkmark & \cellcolor{gray!20}\checkmark & \cellcolor{gray!20}\checkmark & \cellcolor{gray!20}\textbf{10.99} & \cellcolor{gray!20}\textbf{13.99} & \cellcolor{gray!20}\textbf{62.75} & \cellcolor{gray!20}\textbf{30.22} \\
\bottomrule
\end{tabular}
}
\vspace{-0.4cm}
\label{tab:ablation}
\end{table}

\begin{figure*}[th]
    \centering
    \includegraphics[width=0.9\linewidth]{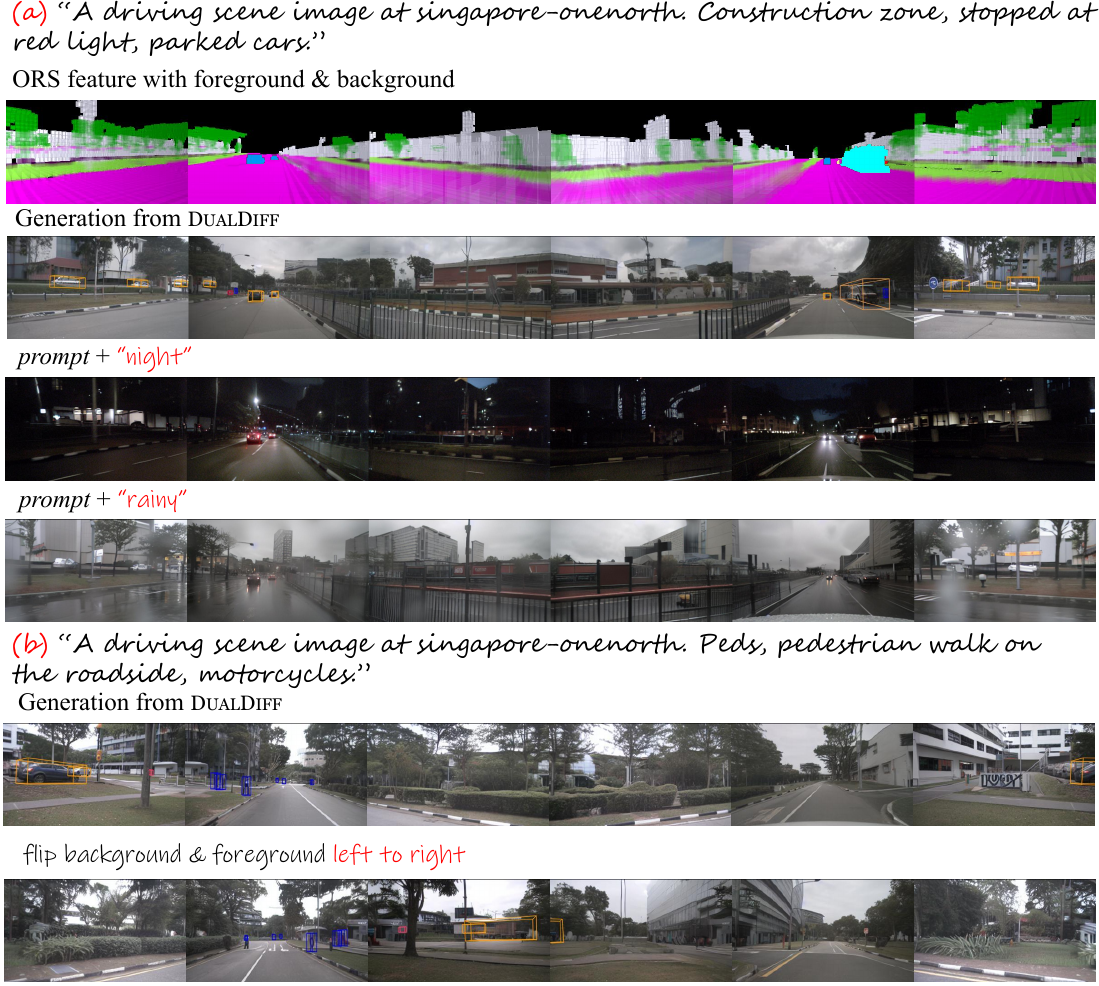}
    \caption{Showcasing Multi-Level Control with DualDiff. We demonstrate scene-level, background-level, and foreground-level control under varying conditions, using scenes from the nuScenes validation set.}
    \label{fig:main_vis}
    \vspace{-0.4cm}
\end{figure*}

\noindent{\textbf{Multi-level Controls.}}
DualDiff utilizes multi-level control conditions to generate accurate street views, organized as follows: 1) Scene Level: Describes high-level attributes such as time, weather, and city, as specified in the scene caption; 2) Background Level: Employs vectorized maps and Occupancy Ray-shape (ORS) background features to control precise background information; 3) Foreground Level: Uses 3D bounding boxes and ORS foreground features to accurately generate foreground objects. As shown in Fig. \ref{fig:main_vis}, DualDiff adapts flexibly to changes at each level, ensuring consistent and realistic scene generation based on the specified conditions.

%% file: main.bbl
\begin{thebibliography}{10}
\providecommand{\url}[1]{#1}
\csname url@samestyle\endcsname
\providecommand{\newblock}{\relax}
\providecommand{\bibinfo}[2]{#2}
\providecommand{\BIBentrySTDinterwordspacing}{\spaceskip=0pt\relax}
\providecommand{\BIBentryALTinterwordstretchfactor}{4}
\providecommand{\BIBentryALTinterwordspacing}{\spaceskip=\fontdimen2\font plus
\BIBentryALTinterwordstretchfactor\fontdimen3\font minus
  \fontdimen4\font\relax}
\providecommand{\BIBforeignlanguage}[2]{{%
\expandafter\ifx\csname l@#1\endcsname\relax
\typeout{** WARNING: IEEEtran.bst: No hyphenation pattern has been}%
\typeout{** loaded for the language `#1'. Using the pattern for}%
\typeout{** the default language instead.}%
\else
\language=\csname l@#1\endcsname
\fi
#2}}
\providecommand{\BIBdecl}{\relax}
\BIBdecl

\bibitem{cui2024survey}
C.~Cui, Y.~Ma, X.~Cao, W.~Ye, Y.~Zhou, K.~Liang, J.~Chen, J.~Lu, Z.~Yang, K.-D.
  Liao \emph{et~al.}, ``A survey on multimodal large language models for
  autonomous driving,'' in \emph{Proceedings of the IEEE/CVF Winter Conference
  on Applications of Computer Vision}, 2024, pp. 958--979.

\bibitem{MARTINEZDIAZ2018275}
\BIBentryALTinterwordspacing
M.~Martínez-Díaz and F.~Soriguera, ``Autonomous vehicles: theoretical and
  practical challenges,'' \emph{Transportation Research Procedia}, vol.~33, pp.
  275--282, 2018, xIII Conference on Transport Engineering, CIT2018. [Online].
  Available:
  \url{https://www.sciencedirect.com/science/article/pii/S2352146518302606}
\BIBentrySTDinterwordspacing

\bibitem{chen2024end}
L.~Chen, P.~Wu, K.~Chitta, B.~Jaeger, A.~Geiger, and H.~Li, ``End-to-end
  autonomous driving: Challenges and frontiers,'' \emph{IEEE Transactions on
  Pattern Analysis and Machine Intelligence}, 2024.

\bibitem{BOSCH201876}
\BIBentryALTinterwordspacing
P.~M. Bösch, F.~Becker, H.~Becker, and K.~W. Axhausen, ``Cost-based analysis
  of autonomous mobility services,'' \emph{Transport Policy}, vol.~64, pp.
  76--91, 2018. [Online]. Available:
  \url{https://www.sciencedirect.com/science/article/pii/S0967070X17300811}
\BIBentrySTDinterwordspacing

\bibitem{10432820}
L.~Szabó and Z.~Weltsch, ``A comprehensive review of existing datasets for
  off-road autonomous vehicles,'' in \emph{2024 IEEE 22nd World Symposium on
  Applied Machine Intelligence and Informatics (SAMI)}, 2024, pp.
  000\,403--000\,410.

\bibitem{rombach2022high}
R.~Rombach, A.~Blattmann, D.~Lorenz, P.~Esser, and B.~Ommer, ``High-resolution
  image synthesis with latent diffusion models,'' in \emph{Proceedings of the
  IEEE/CVF conference on computer vision and pattern recognition}, 2022, pp.
  10\,684--10\,695.

\bibitem{li2023drivingdiffusion}
X.~Li, Y.~Zhang, and X.~Ye, ``Drivingdiffusion: Layout-guided multi-view
  driving scene video generation with latent diffusion model,'' \emph{arXiv
  preprint arXiv:2310.07771}, 2023.

\bibitem{song2020denoising}
J.~Song, C.~Meng, and S.~Ermon, ``Denoising diffusion implicit models,''
  \emph{arXiv preprint arXiv:2010.02502}, 2020.

\bibitem{wen2024panacea}
Y.~Wen, Y.~Zhao, Y.~Liu, F.~Jia, Y.~Wang, C.~Luo, C.~Zhang, T.~Wang, X.~Sun,
  and X.~Zhang, ``Panacea: Panoramic and controllable video generation for
  autonomous driving,'' in \emph{Proceedings of the IEEE/CVF Conference on
  Computer Vision and Pattern Recognition}, 2024, pp. 6902--6912.

\bibitem{gao2023magicdrive}
R.~Gao, K.~Chen, E.~Xie, L.~Hong, Z.~Li, D.-Y. Yeung, and Q.~Xu, ``Magicdrive:
  Street view generation with diverse 3d geometry control,'' in
  \emph{International Conference on Learning Representations}, 2023.

\bibitem{zhang2024perldiff}
J.~Zhang, H.~Sheng, S.~Cai, B.~Deng, Q.~Liang, W.~Li, Y.~Fu, J.~Ye, and S.~Gu,
  ``Perldiff: Controllable street view synthesis using perspective-layout
  diffusion models,'' \emph{arXiv preprint arXiv:2407.06109}, 2024.

\bibitem{wang2024drivewm}
Y.~Wang, J.~He, L.~Fan, H.~Li, Y.~Chen, and Z.~Zhang, ``Driving into the
  future: Multiview visual forecasting and planning with world model for
  autonomous driving,'' in \emph{Proceedings of the IEEE/CVF Conference on
  Computer Vision and Pattern Recognition}, 2024, pp. 14\,749--14\,759.

\bibitem{dualdiff2025}
H.~Li, Z.~Yang, Z.~Qian, G.~Zhao, Y.~Huang, J.~Yu, and L.~Liu, ``Dualdiff:
  Dual-branch diffusion model for autonomous driving with semantic fusion,'' in
  \emph{2025 IEEE International Conference on Robotics and Automation (ICRA)},
  2025, accepted for publication in ICRA 2025.

\bibitem{zhang2023controlnet}
L.~Zhang, A.~Rao, and M.~Agrawala, ``Adding conditional control to
  text-to-image diffusion models,'' in \emph{Proceedings of the IEEE/CVF
  International Conference on Computer Vision}, 2023, pp. 3836--3847.

\bibitem{zhao2024uni}
S.~Zhao, D.~Chen, Y.-C. Chen, J.~Bao, S.~Hao, L.~Yuan, and K.-Y.~K. Wong,
  ``Uni-controlnet: All-in-one control to text-to-image diffusion models,''
  \emph{Advances in Neural Information Processing Systems}, vol.~36, 2024.

\bibitem{wang2023disco}
T.~Wang, L.~Li, K.~Lin, Y.~Zhai, C.-C. Lin, Z.~Yang, H.~Zhang, Z.~Liu, and
  L.~Wang, ``Disco: Disentangled control for realistic human dance
  generation,'' \emph{arXiv preprint arXiv:2307.00040}, 2023.

\bibitem{Singer2022makeavideo}
\BIBentryALTinterwordspacing
U.~Singer, ``Make-a-video: Text-to-video generation without text-video data,''
  2022. [Online]. Available: \url{https://makeavideo.studio/Make-A-Video.pdf}
\BIBentrySTDinterwordspacing

\bibitem{he2023animate}
Y.~He, M.~Xia, H.~Chen, X.~Cun, Y.~Gong, J.~Xing, Y.~Zhang, X.~Wang, C.~Weng,
  Y.~Shan \emph{et~al.}, ``Animate-a-story: Storytelling with
  retrieval-augmented video generation,'' \emph{arXiv preprint
  arXiv:2307.06940}, 2023.

\bibitem{yang2023bevcontrol}
K.~Yang, E.~Ma, J.~Peng, Q.~Guo, D.~Lin, and K.~Yu, ``Bevcontrol: Accurately
  controlling street-view elements with multi-perspective consistency via bev
  sketch layout,'' \emph{arXiv preprint arXiv:2308.01661}, 2023.

\bibitem{wang2023drivedreamer}
X.~Wang, Z.~Zhu, G.~Huang, X.~Chen, J.~Zhu, and J.~Lu, ``Drivedreamer: Towards
  real-world-driven world models for autonomous driving,'' \emph{arXiv preprint
  arXiv:2309.09777}, 2023.

\bibitem{zhao2024drivedreamer2}
G.~Zhao, X.~Wang, Z.~Zhu, X.~Chen, G.~Huang, X.~Bao, and X.~Wang,
  ``Drivedreamer-2: Llm-enhanced world models for diverse driving video
  generation,'' \emph{arXiv preprint arXiv:2403.06845}, 2024.

\bibitem{wu2023tuneavideo}
J.~Z. Wu, Y.~Ge, X.~Wang, S.~W. Lei, Y.~Gu, Y.~Shi, W.~Hsu, Y.~Shan, X.~Qie,
  and M.~Z. Shou, ``Tune-a-video: One-shot tuning of image diffusion models for
  text-to-video generation,'' in \emph{Proceedings of the IEEE/CVF
  International Conference on Computer Vision (ICCV)}, 2023, pp. 7623--7633.

\bibitem{hu2021lora}
E.~J. Hu, Y.~Shen, P.~Wallis, Z.~Allen-Zhu, Y.~Li, S.~Wang, L.~Wang, and
  W.~Chen, ``Lora: Low-rank adaptation of large language models,'' \emph{arXiv
  preprint arXiv:2106.09685}, 2021.

\bibitem{carreira2018quovadisactionrecognition}
\BIBentryALTinterwordspacing
J.~Carreira and A.~Zisserman, ``Quo vadis, action recognition? a new model and
  the kinetics dataset,'' 2018. [Online]. Available:
  \url{https://arxiv.org/abs/1705.07750}
\BIBentrySTDinterwordspacing

\bibitem{whitted2005improved}
T.~Whitted, ``An improved illumination model for shaded display,'' in \emph{ACM
  Siggraph 2005 Courses}, 2005, pp. 4--es.

\bibitem{radford2021clip}
A.~Radford, J.~W. Kim, C.~Hallacy, A.~Ramesh, G.~Goh, S.~Agarwal, G.~Sastry,
  A.~Askell, P.~Mishkin, J.~Clark \emph{et~al.}, ``Learning transferable visual
  models from natural language supervision,'' in \emph{International conference
  on machine learning}.\hskip 1em plus 0.5em minus 0.4em\relax PMLR, 2021, pp.
  8748--8763.

\bibitem{mildenhall2021nerf}
B.~Mildenhall, P.~P. Srinivasan, M.~Tancik, J.~T. Barron, R.~Ramamoorthi, and
  R.~Ng, ``Nerf: Representing scenes as neural radiance fields for view
  synthesis,'' \emph{Communications of the ACM}, vol.~65, no.~1, pp. 99--106,
  2021.

\bibitem{goodfellow2016deep}
I.~Goodfellow, Y.~Bengio, and A.~Courville, \emph{Deep Learning}.\hskip 1em
  plus 0.5em minus 0.4em\relax MIT Press, 2016.

\bibitem{esser2020vqvae}
P.~Esser, R.~Rombach, and B.~Ommer, ``Taming transformers for high-resolution
  image synthesis,'' pp. 12\,873--12\,883, 2021.

\bibitem{vaswani2017attention}
A.~Vaswani, N.~Shazeer, N.~Parmar, J.~Uszkoreit, L.~Jones, A.~N. Gomez,
  L.~Kaiser, and I.~Polosukhin, ``Attention is all you need,'' in
  \emph{Advances in neural information processing systems}, vol.~30, 2017.

\bibitem{dhingra2017gated}
B.~Dhingra, H.~Liu, Z.~Yang, W.~W. Cohen, and R.~Salakhutdinov,
  ``Gated-attention readers for text comprehension,'' in \emph{Proceedings of
  the 55th Annual Meeting of the Association for Computational Linguistics
  (Volume 1: Long Papers)}, 2017, pp. 1832--1846.

\bibitem{zhu2020deformable}
X.~Zhu, W.~Su, L.~Lu, B.~Li, X.~Wang, and J.~Dai, ``Deformable detr: Deformable
  transformers for end-to-end object detection,'' \emph{arXiv preprint
  arXiv:2010.04159}, 2020.

\bibitem{ronneberger2015u}
O.~Ronneberger, P.~Fischer, and T.~Brox, ``U-net: Convolutional networks for
  biomedical image segmentation,'' in \emph{Medical image computing and
  computer-assisted intervention--MICCAI 2015: 18th international conference,
  Munich, Germany, October 5-9, 2015, proceedings, part III 18}.\hskip 1em plus
  0.5em minus 0.4em\relax Springer, 2015, pp. 234--241.

\bibitem{caesar2020nuscenes}
H.~Caesar, V.~Bankiti, A.~H. Lang, S.~Vora, V.~E. Liong, Q.~Xu, A.~Krishnan,
  Y.~Pan, G.~Baldan, and O.~Beijbom, ``nuscenes: A multimodal dataset for
  autonomous driving,'' in \emph{Proceedings of the IEEE/CVF conference on
  computer vision and pattern recognition}, 2020, pp. 11\,621--11\,631.

\bibitem{Sun_2020_waymo}
P.~Sun, H.~Kretzschmar, X.~Dotiwalla, A.~Chouard, V.~Patnaik, P.~Tsui, J.~Guo,
  Y.~Zhou, Y.~Chai, B.~Caine, V.~Vasudevan, W.~Han, J.~Ngiam, H.~Zhao,
  A.~Timofeev, S.~Ettinger, M.~Krivokon, A.~Gao, A.~Joshi, Y.~Zhang, J.~Shlens,
  Z.~Chen, and D.~Anguelov, ``Scalability in perception for autonomous driving:
  Waymo open dataset,'' in \emph{Proceedings of the IEEE/CVF Conference on
  Computer Vision and Pattern Recognition (CVPR)}, June 2020.

\bibitem{heusel2017gans}
M.~Heusel, H.~Ramsauer, T.~Unterthiner, B.~Nessler, and S.~Hochreiter, ``Gans
  trained by a two time-scale update rule converge to a local nash
  equilibrium,'' in \emph{Advances in Neural Information Processing Systems},
  2017.

\bibitem{unterthiner2018fvd}
T.~Unterthiner, B.~Nessler, G.~Heigold, S.~Szedmak, and S.~Hochreiter,
  ``Towards accurate generative models of video: A new metric and challenges,''
  in \emph{Workshop on Challenges and Opportunities for AI in Financial
  Services at NeurIPS}, 2018.

\bibitem{zhou2022cvt}
B.~Zhou and P.~Kr{\"a}henb{\"u}hl, ``Cross-view transformers for real-time
  map-view semantic segmentation,'' in \emph{CVPR}, 2022.

\bibitem{liu2022bevfusion}
Z.~Liu, H.~Tang, A.~Amini, X.~Yang, H.~Mao, D.~Rus, and S.~Han, ``Bevfusion:
  Multi-task multi-sensor fusion with unified bird's-eye view representation,''
  in \emph{IEEE International Conference on Robotics and Automation (ICRA)},
  2023.

\bibitem{li2022bevformer}
Z.~Li, W.~Wang, H.~Li, E.~Xie, C.~Sima, T.~Lu, Y.~Qiao, and J.~Dai,
  ``Bevformer: Learning bird’s-eye-view representation from multi-camera
  images via spatiotemporal transformers,'' in \emph{European conference on
  computer vision}.\hskip 1em plus 0.5em minus 0.4em\relax Springer, 2022, pp.
  1--18.

\bibitem{coda2024}
\BIBentryALTinterwordspacing
{CODA Dataset}, ``w-coda 2024 track 2,'' 2024, accessed: 2024-01-07. [Online].
  Available: \url{https://coda-dataset.github.io/w-coda2024/track2/}
\BIBentrySTDinterwordspacing

\bibitem{wang2023streampetr}
S.~Wang, Y.~Liu, T.~Wang, Y.~Li, and X.~Zhang, ``Exploring object-centric
  temporal modeling for efficient multi-view 3d object detection,'' in
  \emph{Proceedings of the IEEE/CVF International Conference on Computer
  Vision}, 2023, pp. 3621--3631.

\bibitem{zhao2024unipc}
W.~Zhao, L.~Bai, Y.~Rao, J.~Zhou, and J.~Lu, ``Unipc: A unified
  predictor-corrector framework for fast sampling of diffusion models,''
  \emph{Advances in Neural Information Processing Systems}, vol.~36, 2024.

\bibitem{ho2022classifier}
J.~Ho, X.~Chen, A.~Srinivas, and et~al., ``Classifier-free diffusion
  guidance,'' in \emph{NeurIPS 2022}, 2022.

\bibitem{swerdlow2024bevgen}
A.~Swerdlow, R.~Xu, and B.~Zhou, ``Street-view image generation from a
  bird's-eye view layout,'' \emph{IEEE Robotics and Automation Letters}, 2024.

\bibitem{kim2021drivegan}
S.~W. Kim, J.~Philion, A.~Torralba, and S.~Fidler, ``Drivegan: Towards a
  controllable high-quality neural simulation,'' in \emph{Proceedings of the
  IEEE/CVF Conference on Computer Vision and Pattern Recognition}, 2021, pp.
  5820--5829.

\bibitem{zheng2025genad}
W.~Zheng, R.~Song, X.~Guo, C.~Zhang, and L.~Chen, ``Genad: Generative
  end-to-end autonomous driving,'' in \emph{European Conference on Computer
  Vision}.\hskip 1em plus 0.5em minus 0.4em\relax Springer, 2025, pp. 87--104.

\bibitem{li2025drivingdiffusion}
X.~Li, Y.~Zhang, and X.~Ye, ``Drivingdiffusion: Layout-guided multi-view
  driving scenarios video generation with latent diffusion model,'' in
  \emph{European Conference on Computer Vision}.\hskip 1em plus 0.5em minus
  0.4em\relax Springer, 2025, pp. 469--485.

\bibitem{zhou2024simgen}
Y.~Zhou, M.~Simon, Z.~Peng, S.~Mo, H.~Zhu, M.~Guo, and B.~Zhou, ``Simgen:
  Simulator-conditioned driving scene generation,'' \emph{arXiv preprint
  arXiv:2406.09386}, 2024.

\bibitem{ma2024delphi}
E.~Ma, L.~Zhou, T.~Tang, Z.~Zhang, D.~Han, J.~Jiang, K.~Zhan, P.~Jia, X.~Lang,
  H.~Sun \emph{et~al.}, ``Unleashing generalization of end-to-end autonomous
  driving with controllable long video generation,'' \emph{arXiv preprint
  arXiv:2406.01349}, 2024.

\bibitem{zheng2023layoutdiffusion}
G.~Zheng, X.~Zhou, X.~Li, Z.~Qi, Y.~Shan, and X.~Li, ``Layoutdiffusion:
  Controllable diffusion model for layout-to-image generation,'' in
  \emph{Proceedings of the IEEE/CVF Conference on Computer Vision and Pattern
  Recognition}, 2023, pp. 22\,490--22\,499.

\bibitem{li2023gligen}
Y.~Li, H.~Liu, Q.~Wu, F.~Mu, J.~Yang, J.~Gao, C.~Li, and Y.~J. Lee, ``Gligen:
  Open-set grounded text-to-image generation,'' in \emph{Proceedings of the
  IEEE/CVF Conference on Computer Vision and Pattern Recognition}, 2023, pp.
  22\,511--22\,521.

\bibitem{li2023blip2}
J.~Li, D.~Li, J.~Gao \emph{et~al.}, ``Blip-2: Bootstrapping language-image
  pretraining with frozen vision-language models,'' in \emph{Proceedings of the
  IEEE/CVF Conference on Computer Vision and Pattern Recognition (CVPR)}, 2023.

\end{thebibliography}
